\documentclass{article}

\usepackage{PRIMEarxiv}

\usepackage[utf8]{inputenc} % allow utf-8 input
\usepackage[T1]{fontenc}    % use 8-bit T1 fonts
\usepackage{amssymb}
\usepackage{amsmath}
\usepackage{amsfonts}
\usepackage{booktabs}       % professional-quality tables
\usepackage{nicefrac}
\usepackage{microtype}      % microtypography
\usepackage{graphicx}
\usepackage{longtable}
\usepackage{array}
\usepackage{ragged2e}
\usepackage{fancyhdr}       % header
\usepackage[numbers,sort&compress]{natbib}
\usepackage[hyphens]{url}  % let the long GitHub URL break at hyphens
\usepackage{hyperref}       % hyperlinks (load last)
\hypersetup{colorlinks=true, linkcolor=blue, citecolor=blue, urlcolor=blue}

\title{When Synthetic Users Fail: A Cross-Domain Benchmark of\\
LLM-Simulated Human Survey Responses
\thanks{This manuscript is currently under peer review. Code and data are publicly
available at: \url{https://github.com/ZihanChen1995/when-synthetic-users-fail-a-cross-domain-benchmark-of-llm-simulated-human-survey-responses}}}

\author{
  Zihan Chen \\
  Stevens Institute of Technology \\
  Hoboken, NJ, USA \\
  \texttt{zchen61@stevens.edu} \\
  \And
  Di Zhu \\
  Stevens Institute of Technology \\
  Hoboken, NJ, USA \\
  \texttt{dzhu1@stevens.edu} \\
  \And
  Lei Nico Zheng \\
  University of Massachusetts Boston \\
  Boston, MA, USA \\
  \texttt{lei.zheng@umb.edu} \\
}

\begin{document}
\maketitle
%% \thanks consumed footnote 1; reset so numbered footnotes still start at 1.
\setcounter{footnote}{0}

\begin{abstract}
Large language models (LLMs) are increasingly used as synthetic
users, stand-ins for human respondents whose simulated answers feed product,
policy, and market decisions. We ask when this substitution is valid and when it
fails, and package the answer as an evaluation framework for intelligent
synthetic-user systems. A single protocol, run across four models spanning two families and an
8B-to-frontier capability range, is applied to two independent domains of
real human-response data: U.S.\ general social attitudes (General Social Survey)
and cross-cultural values (World Values Survey). Every model is
benchmarked against a suite of non-LLM baselines fit on held-out human data. Under
demographic prompting and the survey-simulation protocols we test, two failures
replicate across both domains, all four models, and both families. First, at the
individual level no LLM beats even the strongest baseline; on cross-cultural values
every model falls well below it, and the gap survives distance-aware and proper
scoring. Second, models systematically \emph{over-determine} demographics, treating
identity as far more predictive of attitudes than it is among real people, a
distortion present for nearly every question--group combination and robust to a
coding-invariant measure. Neither failure is remedied by a larger, more capable
model. A decision-impact analysis shows why this matters in practice: on a segment-targeting
task the models inflate between-segment gaps two to fourfold, would direct a team
to the wrong segment in half of U.S.\ and most cross-cultural cases, and
manufacture segment splits that do not exist in real people. We publicly
release the cross-domain benchmark and validation framework, including all code and
data, so that teams can determine in advance when synthetic-user evidence is safe for decision support and when it is not.
\end{abstract}


%% ======================================================================
\section{Introduction}
\label{sec:intro}

Running a human study is slow and expensive, so a fast-growing body of work
proposes replacing human participants with large language models (LLMs) prompted
to behave as \emph{synthetic users}, using them to pilot surveys, pre-test
questionnaire items, estimate opinion distributions, and stand in for study
participants. The appeal is obvious: a survey that takes months and thousands of
dollars to field can, in principle, be approximated in minutes at the cost of a few
API calls.

These simulated respondents are increasingly wired into \emph{decision-support}
pipelines. Product teams query synthetic personas to prioritize features, marketers
use them to pre-test messaging and estimate segment-level demand through ``silicon
samples''~\cite{sarstedt2024silicon,li2024frontiers}, and policy and organizational
analysts use them to gauge how a population might react before committing resources.
In each case an intelligent system is asked to manufacture evidence that a human
sample would normally provide, and a downstream decision is made on that evidence.
Viewed this way, a synthetic user is an evidence-producing component of an
intelligent decision-support system, and, like any such component, it should be
verified and validated against its intended use before it is
trusted~\cite{shmueli2011predictive,okeefe1993expert}. This is precisely the
setting where an unnoticed failure is costly: if the synthetic users are wrong in a
structured way, the resulting decisions inherit that error at scale\footnote{``Fail'' is scoped throughout to
\emph{demographic prompting under survey-simulation protocols}: the specific elicitation setting we test. We make no claim about synthetic users produced by other means (e.g.\ fine-tuning or richer persona construction), which the
protocol is designed to evaluate in turn.}. A practical evaluation
framework, one that tells a team \emph{in advance} whether synthetic-user evidence
is trustworthy for the decision at hand, is therefore as important as the
simulation method itself. Building that framework is the goal of this paper.

The central unresolved question is not whether LLMs can produce
survey-\emph{shaped} output (they trivially can) but \emph{when the substitution
is valid and when it fails}. The evidence so far is conflicting: some studies
report encouraging aggregate alignment~\cite{argyle2023}, while others warn that
substitution is unsafe~\cite{bisbee2024,gao2025caution}. Prior work has also largely
answered the question one domain at a time (political surveys, or cultural values,
or a single national survey), often reporting a single aggregate similarity number
that can look reassuring even when the underlying simulation is poor. Two questions are therefore
under-answered. (i) Is a synthetic user actually more informative than the
trivial thing it is supposed to replace, a lookup of ``what do people with these
demographics usually answer''? (ii) Do conclusions about validity \emph{transfer
across domains}, or is each positive result an artifact of one survey, one model,
or one prompt?

We address both with a deliberately unified design. We fix one protocol (the
same models, prompts, decoding, sampling, and metrics) and apply it to two
independent domains of \emph{real} human-response ground truth:

\begin{itemize}
  \item \textbf{U.S.\ general social attitudes.} The General Social Survey (GSS),
    2016--2024 waves, 10 attitude questions spanning wellbeing, social trust,
    moral/policy positions, gender-role attitudes, and institutional confidence.
  \item \textbf{Cross-cultural values.} The World Values Survey (WVS) Wave~7, 63
    countries, 16 ordinal value questions spanning economic values, corruption
    perceptions, immigration attitudes, views of science, religiosity, and moral
    justifiability.
\end{itemize}

Across these domains we evaluate four models from two families across an
8B-to-frontier capability range, under two prompt formats (a single-answer prompt
and a probability-distribution prompt). The design choice that makes
our individual-level results interpretable, and which much prior work
omits, is an explicit \textbf{naive demographic baseline}: for each question we
fit the conditional distribution of human answers given demographics on a
held-out portion of the real data, and score every LLM against it on the same
respondents. Without this yardstick, ``the model predicts individuals with
$X\%$ accuracy'' is uninterpretable, because individual survey answers are not a
deterministic function of demographics; there is an irreducible ceiling that a
trivial predictor already captures.

\paragraph{Research questions}
We formalize the study as five research questions. The first three ascend the levels
of analysis at which synthetic users are actually deployed, from the individual
respondent (RQ1), through the population aggregate (RQ2), to the demographic
structure of opinion (RQ3); the final two then ask whether those answers are stable
to modelling choices (RQ4) and whether they transfer across domains (RQ5).
\begin{enumerate}
  \item[\textbf{RQ1}] \emph{Individual fidelity vs.\ a trivial baseline.} Do LLM
    synthetic users predict individual human answers more accurately than a
    naive demographic-conditional predictor?
  \item[\textbf{RQ2}] \emph{Aggregate fidelity.} Do LLM synthetic users reproduce
    the population-level distribution of human answers?
  \item[\textbf{RQ3}] \emph{Subgroup structure.} Do LLMs represent the
    demographic structure of attitudes faithfully, or do they distort how
    predictive demographics are of answers?
  \item[\textbf{RQ4}] \emph{Stability and capability.} Are these behaviors stable
    across model family, model capability, and output format?
  \item[\textbf{RQ5}] \emph{Cross-domain transfer.} Do the answers to RQ1--RQ4
    hold in both domains, or are they domain-specific?
\end{enumerate}

\paragraph{Headline findings}
Our central result is that, under demographic prompting and the survey-simulation
protocols we test, LLM synthetic users exhibit two distinct failures that replicate
across \emph{both} domains, \emph{all four} models, and \emph{both} families. The
first is a \textbf{lack of individual-level advantage.} As yardsticks we
build a suite of non-LLM predictors, from a ``demographic lookup table'' (for any
set of demographics, it returns the answer most common among real people with
those demographics) up to learned demographic models (logistic regression, random
forest). On GSS (U.S.\ attitudes) every LLM, at best, only ties the lookup table
and trails the learned baseline; on WVS (cross-cultural values) every model is
\emph{11 to 22 percentage points less accurate} than the baseline. The gap is not
an artifact of exact-match scoring: it persists under distance-aware metrics that
give partial credit for near-misses on ordinal scales, and under proper scoring of
the distribution outputs, where the model assigns \emph{lower} probability to the
true human answer than the baseline. In short, an LLM asked to role-play an
individual adds no information beyond what the demographics alone already
imply, and for value questions it is substantially worse. The second failure is
\textbf{demographic over-determination}, or stereotyping. We measure how strongly a
demographic attribute predicts a person's answer (the share of the variation in answers that is
explained by which group the person is in) and compare that share for real humans
versus for the model. Models consistently exaggerate it. For U.S.\ political
leaning and confidence in banks, for example, a person's politics explains only
about $1.5\%$ of the variation in real answers, yet the model behaves as though it
explains up to roughly $67\%$. This exaggeration is present, and statistically
distinguishable from chance, for nearly every question--group combination in both
domains, and it replicates under a coding-invariant association measure that makes
no numeric assumption about the answer scale. Crucially, \textbf{neither failure is
fixed by using a bigger, more capable model}: the frontier models stereotype at
least as strongly as the small 8B model, and often more.

\paragraph{Contributions}
Taken together, our contributions are both methodological and practical. On the
\emph{methodological and empirical} side we make four. First, we build a compact,
reproducible \textbf{cross-domain benchmark} for LLM synthetic users, spanning U.S.\
social attitudes and 63-country values under a single protocol and packaged as a
reusable, publicly released evaluation toolkit (data-build, inference, and analysis
scripts, with all prompts, parser rules, and baselines specified in the appendices). Second, we develop
a \textbf{baseline-anchored evaluation}: we show that the standard individual-accuracy
number is uninterpretable without non-LLM baselines, and that against a suite of them
(question-marginal, demographic lookup, and learned demographic models) LLMs show no
individual-level advantage, a result we confirm with distance-aware metrics for
ordinal scales and proper scoring (log-loss, Brier) for distribution outputs, with
paired-bootstrap confidence intervals throughout. Third, we introduce a
\textbf{stereotyping index}, a single, bounded number that compares how predictive a
demographic attribute is of the answer in the model versus in real people, reported
with confidence intervals and paired with a coding-invariant companion measure
(Cram\'er's~V) that requires no numeric coding of answers; it reframes the common
worry that ``LLMs flatten minorities'' into its measurable opposite for
identity-linked attitudes, namely that LLMs \emph{over-associate} demographics with
answers, treating identity as more decisive than it really is. Fourth, we provide a
\textbf{decision-impact analysis} that carries the over-determination finding through
to the decision a synthetic user is meant to support: on the canonical
segment-targeting task, we quantify how far the models inflate between-segment gaps
(two to fourfold), how often they would send a team to the wrong segment
($50$--$72\%$ of cases), and how often they manufacture a segment split that does not
exist in real people (up to $41\%$ of cases on cross-cultural values).

On the \emph{practical, applied-intelligent-systems} side, these pieces combine into
our main deliverable: a \textbf{validation framework for intelligent synthetic-user
systems} used in decision support. It is not merely a benchmark of model outputs but a
verification-and-validation procedure a practitioner runs \emph{before} deployment,
one that reports individual and aggregate fidelity separately, always against non-LLM
baselines; reports subgroup determinism and its decision impact; reports
invalid-output rates; and never assumes a larger model is a safer synthetic user. The
framework is the deliverable; the negative findings are the evidence that a framework
of this kind is necessary.

The remainder of the paper reviews related work
(Section~\ref{sec:related}), details the data, protocol, and metrics
(Section~\ref{sec:method}), presents results by research question
(Section~\ref{sec:results}), and discusses implications and limitations
(Sections~\ref{sec:discussion}--\ref{sec:conclusion}).

%% ======================================================================
\section{Related Work}
\label{sec:related}

Our study sits at the intersection of six literatures. We review each in turn,
then state the gap that motivates a baseline-anchored, cross-domain validation
framework.

\subsection{LLMs as simulated survey respondents and participants}
\label{sec:rw-respondents}
A fast-growing line of work treats LLMs as programmable populations, prompting them
with personas or demographics to predict opinions and behaviors. The optimistic
anchor is \citet{argyle2023}, who show that demographically conditioned prompts can
reproduce aggregate patterns in U.S.\ political survey data, launching the
``silicon sample'' idea; related work extends the substitution to behavioral
experiments~\cite{aher2023using}, economic agents~\cite{horton2023homo}, and the
generation of synthetic research data in HCI and product
settings~\cite{hamalainen2023evaluating}.
A skeptical counter-current quickly followed. \citet{bisbee2024} show that
LLM-generated survey responses can misrepresent population and subgroup variation;
\citet{santurkar2023whose} find that model ``opinions'' align more with some
demographic groups than others rather than neutrally representing a population; and
\citet{dominguezolmedo2024questioning} show that model survey answers are sensitive
to ordering and labeling, and so should not be read as direct measurements of
population beliefs. Position pieces in cognitive science and psychology sharpen the
caution: replacing human participants risks confusing linguistic plausibility with
validity~\cite{dillion2023can,gao2025caution,lin2025six}. The practical problem for
this paper is precisely this gap between survey-shaped output and valid evidence.

\subsection{Survey-simulation benchmarks and cross-cultural value modeling}
\label{sec:rw-benchmarks}
A parallel effort builds benchmarks that make survey and cross-cultural value
simulation measurable at scale. WorldValuesBench derives a large demographic-context-to-answer
benchmark from the World Values Survey~\cite{worldvaluesbench2024}; SocioBench
evaluates LLMs as survey-response predictors across countries and sociological
domains~\cite{sociobench2025}; and distribution-specialization work fine-tunes
models to match country-level response distributions, improving group-level
prediction while still struggling to generalize~\cite{cao2025}. One of
our two domains reuses the WorldValuesBench question set and scales for
comparability. Our work differs from these benchmarks on three axes that matter
for a validity claim. First, \emph{reference point}: WorldValuesBench and
SocioBench score model accuracy against the human answer, but not against a non-LLM
predictor of that answer, so a reported accuracy cannot be read as evidence that
the simulation adds information; our demographic-lookup and learned baselines
supply exactly that missing reference, and against it the individual-level
advantage disappears. Second, \emph{level of analysis}: these benchmarks (and the
distribution-specialization work of \citet{cao2025}, which improves
group-level fit through fine-tuning) target aggregate or country-level
distributions, which our RQ2 confirms models reproduce reasonably; we show that
aggregate success coexists with individual-level failure and with a subgroup
distortion the aggregate metric cannot see. Third, \emph{cross-domain transfer}:
each benchmark fixes a single survey instrument, so a positive or negative result
could be an artifact of that instrument; by holding one protocol fixed across a
U.S.\ attitude survey and a 63-country value survey we can separate properties of
LLM simulation from properties of a dataset. In short, we complement calibrated
value and persona benchmarks by asking whether simulated users add individual-level
information beyond a demographic baseline, whether the demographic structure of
answers is faithful, and whether the answer transfers across domains.

\subsection{Persona prompting, silicon samples, and prompt artifacts}
\label{sec:rw-persona}
Persona and demographic prompting is the dominant mechanism for constructing
synthetic users, and it is used well beyond political surveys: in marketing and
consumer research, where ``silicon samples'' are proposed as low-cost respondent
pools~\cite{sarstedt2024silicon}, and in automated perceptual analysis for product
decisions~\cite{li2024frontiers}. Crucially, persona assignment is not a neutral
conditioning operation. It can amplify toxicity and stereotypes~\cite{deshpande2023toxicity},
the surface architecture of a prompt can induce methodological
artifacts~\cite{brucks2025prompt}, sociodemographic persona formulations
materially change model behavior~\cite{lutz2025prompt}, and persona generation
itself carries quality and validity pitfalls~\cite{li2025persona}. This literature
motivates treating prompt-surface stability and demographic over-determination
(our RQ3--RQ4) as first-order validity conditions rather than minor robustness
checks.

\subsection{Social simulation and LLM agents}
\label{sec:rw-agents}
Synthetic survey respondents are one instance of a broader movement to simulate
humans with LLMs. Generative agents endow LLMs with memory, reflection, and
planning to produce believable individual and social behavior~\cite{park2023generative},
building on earlier populated prototypes for social computing
systems~\cite{park2022social}; recent surveys map the field from individual agents
to whole-society simulation~\cite{mou2026from} and assess the promise and epistemic
risks of LLMs for computational social science~\cite{ziems2024can}. This work
mostly evaluates believability or downstream system behavior; we contribute a
quantitative, human-anchored validity benchmark for one high-stakes class of these
simulations.

\subsection{Algorithmic stereotyping, fairness, and subgroup validity}
\label{sec:rw-fairness}
Our over-determination finding connects to a long line of work on social bias in
learned representations: human-like biases in embeddings~\cite{caliskan2017semantics,bolukbasi2016man},
stereotyped associations in generation~\cite{sheng2019woman}, and representational
harms of large models~\cite{bender2021dangers}. The fairness literature further
establishes that aggregate performance can mask subgroup harm: intersectional
accuracy disparities~\cite{buolamwini2018gender}, group-conditioned error
criteria~\cite{hardt2016equality}, hand-built bias benchmarks for constrained-choice
QA~\cite{parrish2022bbq}, and the reminder that fairness must be assessed within the
sociotechnical context of use~\cite{selbst2019fairness}. Our stereotyping index
adapts this subgroup lens but differs in target: rather than asking whether a model
attaches a negative label to a group, we compare \emph{how predictive} demographics
are of an answer in the model versus in real humans, turning stereotyping into a
human-comparable validity metric.

\subsection{Decision-support validity and expert-system validation}
\label{sec:rw-dss}
For an applied-AI venue, the decisive framing is that synthetic users are
evidence-producing components inside intelligent decision-support systems, and such
components require verification and validation against intended use. Information-systems
research treats predictive validity (out-of-sample performance against
baselines) as a first-class evaluation goal distinct from explanatory
fit~\cite{shmueli2011predictive}; the decision-support and expert-systems tradition
has long emphasized verification, validation, and evaluation of intelligent
artifacts~\cite{arnott2005critical,okeefe1993expert,saibene2021expert}. Multi-metric
reporting norms from ML evaluation reinforce the same discipline: holistic
evaluation across accuracy, calibration, robustness, and
fairness~\cite{liang2023holistic}, behavioral testing beyond aggregate
accuracy~\cite{ribeiro2020checklist}, broad capability benchmarks~\cite{srivastava2023bigbench},
and standardized documentation of models and datasets~\cite{mitchell2019modelcards,gebru2021datasheets}.
Finally, survey methodology's total-survey-error framework distinguishes
representation from measurement error~\cite{groves2010total}, which is exactly why
individual, aggregate, and subgroup validity are separate questions: an LLM
synthetic user introduces a new error source that can look benign in aggregate
while failing at the individual or subgroup level.

\subsection{What is missing}
\label{sec:rw-gap}
This prior work is strong but fragmented: each study typically fixes one domain, one
or two models, and one evaluation lens, which makes it hard to tell whether a
reported failure (or success) is a property of LLM simulation or of the specific
setup. Two gaps matter for practice. First, individual-level results are usually
reported without \emph{non-LLM baselines}, so it is unclear whether the model beats
a demographic lookup, let alone a learned demographic model~\cite{shmueli2011predictive}, and
whether any apparent gap is merely an artifact of harsh exact-match scoring on
ordinal scales. Second, subgroup analyses typically ask whether groups are
``flattened,'' but do not quantify, on a human-comparable scale, how much
demographics \emph{should} explain versus how much the model makes them explain. We
close both gaps by anchoring individual fidelity to a baseline suite scored with
distance-aware and proper-scoring metrics, and by quantifying subgroup determinism with
both a variance-based and a coding-invariant association measure. Holding
one protocol fixed across two domains and four models, we then test whether the conclusions
transfer.

%% ======================================================================
\section{Data, Protocol, and Metrics}
\label{sec:method}

\subsection{Domains and scope}
We use two public human-response datasets and state the scope we actually use;
all claims are restricted to that scope.

\textbf{GSS (U.S.\ social attitudes).} We use the GSS 1972--2024 cumulative file
restricted to the \textbf{2016--2024} waves. Recent waves give large
per-subgroup cells and partially mitigate pretraining contamination relative to
decades-old waves. After removing respondents with missing values on any prompt
demographic, the in-scope pool is 14{,}704 respondents and 85{,}898
(respondent, question) pairs over 10 questions. Demographics used in the prompt:
age, sex, race, highest degree, region, political views, party identification.

\textbf{WVS (cross-cultural values).} We use WVS Wave~7 (inverted CSV, v6.0). We
retain the 16 value questions from the WorldValuesBench probe
set~\cite{worldvaluesbench2024} that are present in this release; question wordings
and ordinal scales follow WorldValuesBench for comparability. After demographic filtering the in-scope pool is 91{,}774
respondents across 63 countries and 1{,}426{,}473 (respondent, question) pairs.
Demographics: age group, sex, education level, settlement type (urban/rural), and
country.

\subsection{Sampling for measurable distributions}
\label{sec:sampling}
Our headline analyses need a reliable estimate of the human answer
\emph{distribution} within each (question, group) cell, which requires enough
people in each cell. We therefore sample by cell rather than drawing respondents at
random. For each domain we draw about $100$ respondents per question (GSS: 993
rows; WVS: 1{,}458 rows), allocated so that the main grouping is well covered
(GSS: degree $\times$ race; WVS: country). Importantly, the human answer
distributions and the demographic baseline are always estimated from the full dataset, never from this small sample, so the human side of every
comparison rests on large, stable numbers.

\subsection{Models and prompting}
We evaluate four chat-style models through a single, common interface: Claude
Haiku~4.5 and Claude Sonnet~4.6 (the \emph{closed} family: proprietary, accessed
via an API) and Llama-3.1-8B and Llama-3.3-70B (the \emph{open} family: open-weight
and publicly downloadable). Together these span two independent model families and
a wide capability range, from a small 8B model to frontier-scale systems. Each sampled respondent is run under two prompt
formats and two independent generation runs (``seeds,'' which differ only in the
model's internal randomness, letting us gauge run-to-run noise):
\begin{itemize}
  \item \textbf{Style~A, the single-answer prompt.} The model is shown the
    demographic profile and question and must return one answer (an option letter
    for short-labelled questions, or the scale number for numeric scales).
  \item \textbf{Style~C, the distribution prompt.} The model must return a
    probability for each answer option, expressed as JSON.
\end{itemize}
We use these two names (Style~A / single-answer prompt, Style~C / distribution
prompt) interchangeably throughout. The model's internal randomness is held at a
fixed setting across models; the two seeds probe run-to-run stability. All calls
use the same output parser, which flags any response it cannot read as a valid
answer as \emph{invalid/refused}.

\subsection{Non-LLM baselines}
\label{sec:baselines}
The individual-accuracy number is only interpretable against what a non-LLM
predictor achieves on the same people, so we build a suite of four non-LLM baselines. All are fit on a held-out 50\% split of the in-scope
pool (assigned by a hash of the respondent id) and evaluated on the evaluation-sample rows,
which come from the other split, so no baseline is ever fit and scored on the same
respondents. The four baselines, from simplest to most sophisticated, are as follows
(full formulae and hyperparameters are in Appendix~\ref{app:baselines}):
\begin{itemize}
  \item \textbf{Question marginal.} The single most common answer to the question
    among the fit humans (ignores demographics entirely).
  \item \textbf{Demographic lookup} (our primary yardstick). A predictor of the
    conditional answer distribution given demographics, which backs off from a fine
    demographic cell to a coarse cell to the question marginal, so it always yields a
    prediction. For WVS the
    finest cell begins with country; the coarse cell is country alone, which we also
    report separately as a \emph{country-only} baseline.
  \item \textbf{Multinomial logistic regression} on the one-hot-encoded prompt
    demographics, fit per question.
  \item \textbf{Random forest} on the same features (300 trees), a nonlinear
    learned baseline.
\end{itemize}
The demographic lookup and logistic model also emit a full predicted
\emph{distribution} over answers, which we use as the reference for the proper-scoring
comparison against the model's distribution outputs (Section~\ref{sec:metrics}).
These baselines are the yardsticks for RQ1: following the predictive-validity
tradition in information systems~\cite{shmueli2011predictive}, a synthetic user
earns its place only by beating a simpler predictor on held-out human data.

\subsection{Metrics}
\label{sec:metrics}
We use four families of metric, one per research question. Reporting individual,
aggregate, and subgroup fidelity separately mirrors the total-survey-error
distinction between measurement and representation error~\cite{groves2010total} and
the multi-metric evaluation norms now standard for language
models~\cite{liang2023holistic,ribeiro2020checklist}: a single aggregate number
hides exactly the failures we care about. Each metric is stated in plain terms
first, then defined precisely. Every headline comparison against a baseline carries
a \textbf{paired-bootstrap 95\% confidence interval}: we resample the evaluation-sample
rows $500$ times, scoring model and baseline on the \emph{same} rows in each
resample, and report the interval of the difference, so the reader can tell a real
margin from sampling noise.

\textbf{Individual fidelity (RQ1): can the model guess one person's answer?} We
report \emph{accuracy} (the fraction of individuals whose exact answer the model
predicts correctly) next to each baseline's accuracy on the same people. For the
distribution prompt (Style~C), the model's single ``guess'' is the option it
assigned the highest probability (the arg-max of its distribution). Exact-match
accuracy is deliberately harsh on long ordinal scales, so for the ordinal questions
we also report two \emph{distance-aware} metrics that give partial credit for being
close. The \emph{mean absolute error} (MAE) is the average gap, in scale points,
between the single predicted answer and the true one. The \emph{earth-mover's
distance} (EMD) uses the whole predicted distribution from the distribution prompt:
it is the expected number of scale steps needed to move the predicted probability
mass onto the true answer, and so rewards a distribution that concentrates near the
truth even when its mode is wrong. Both are reported for the model and for the
baseline on the same rows.

\textbf{Aggregate fidelity (RQ2): does the model reproduce the group's answer
\emph{spread}?} Rather than any one person, we compare the overall \emph{share} of
people choosing each option (the answer distribution) between the model and real
humans. We summarize the gap between two distributions with the
\emph{Jensen--Shannon (JS) divergence}, a standard $0$-to-$1$ measure of how
different two probability distributions are; $0$ means identical, and larger means
further apart, so lower is better. Because we sample by cell rather than at random
(Section~\ref{sec:sampling}), the raw pooled model distribution reflects our
sampling design, not the human population. We therefore also report a
\emph{population-reweighted} JS, in which each cell's model distribution is
weighted back to that cell's share of the real population before comparing to the
true (full-table) human distribution. If the two versions agree, the
aggregate-fidelity result is not an artifact of the sampling design.

For the distribution prompt (Style~C), arg-max accuracy discards most of the
predicted distribution, so we additionally score the \emph{whole} distribution with
two proper scoring rules evaluated at the true human answer: the \emph{log-loss}
(negative log-likelihood the model places on the true answer) and the \emph{Brier
score} (squared error of the predicted distribution against the one-hot true
answer). Lower is better, and both are reported against the baseline's conditional
distribution on the same rows; a proper score is the right way to ask whether the
distribution puts mass where the human actually answered.

\textbf{Subgroup determinism (RQ3): does the model treat identity as destiny?}
Fairness research has long argued that aggregate performance can hide subgroup
harm and must be assessed group by group~\cite{buolamwini2018gender,hardt2016equality}.
We adopt that lens but ask a sharper, human-anchored question: how strongly a
demographic attribute (say, political leaning) \emph{predicts} the answer, for real
people and for the model. The natural measure is the share
of the total variation in answers that is explained by which group a person
belongs to. Statisticians call this $\eta^2$ (``eta-squared''), the ratio of the
variation \emph{between} groups to the total variation (between plus within
groups); it runs from $0$ (the attribute tells you nothing about the answer) to
$1$ (the attribute determines the answer completely). We compute $\eta^2$ the same
way for humans and for the model. Our headline measure, the \textbf{stereotyping
index}, is simply the difference,
\[
  \Delta\eta^2 \;=\; \eta^2_{\text{model}} - \eta^2_{\text{human}},
\]
which is positive when the model treats a demographic attribute as \emph{more}
predictive of the answer than it actually is among real people, that is, when the
model stereotypes. Because it is a difference of two bounded quantities computed
identically, it is not distorted by which prompt style produced the model's
answers, and it captures the full spread within each group rather than only the
group averages. We attach a $95\%$ confidence interval to each value by
resampling respondents $500$ times (a bootstrap), so we can tell a real effect
from sampling noise.

$\eta^2$ treats the answer codes as numbers, which is natural for the ordinal
questions but would be an arbitrary choice for nominal ones. To guard against the
objection that the finding depends on how answers are coded, we add a
\emph{coding-invariant} companion measure that uses no numeric codes at all:
Cram\'er's~V, the standard association strength between two categorical variables
(here, group membership and the chosen answer), computed identically for humans
and for the model from the group$\times$answer contingency table. Its stereotyping
analogue is $\Delta V = V_{\text{model}}-V_{\text{human}}$, again with a bootstrap
$95\%$ CI. Because it is purely nominal, we can report it for \emph{all} questions,
including the nominal GSS items that the $\eta^2$ index omits. If both the numeric
$\Delta\eta^2$ and the coding-free $\Delta V$ are positive, over-determination is
not an artifact of the answer coding.

\textbf{Decision impact: what does over-determination cost a decision that acts on
it?} The stereotyping index measures the distortion in the model's world; a
decision-support venue needs to know what that distortion does to a decision. We
therefore translate it into the terms of the single most common use of a synthetic
user in practice: \emph{segment targeting}. A product, marketing, or policy team
asks which demographic segment is most extreme on an attitude, and how large the
between-segment gap is, in order to target the top segment, size a niche, or claim
``this attitude splits along <segment>.'' For each (question, demographic axis) we
compute, identically for the model and for real humans, the \emph{segment share}
$p_g$ (the fraction of segment $g$ on the high end of the answer scale), and read
three quantities a targeting decision keys on: the \emph{between-segment gap}
$\max_g p_g - \min_g p_g$; the \emph{target segment} $\arg\max_g p_g$; and, when
the model's target differs from the human target, the \emph{targeting cost} (the
attitude a team forgoes, in real human share, by acting on the model's segment
instead of the true one). From these we report (i) the \emph{gap-inflation factor}
$\text{gap}_{\text{model}}/\text{gap}_{\text{human}}$, how much the model
exaggerates the between-segment difference; (ii) the \emph{wrong-target rate}, the
fraction of (question, axis) pairs where the model would send a team to the wrong
segment; and (iii) the \emph{spurious-split rate}, the fraction where humans show a
negligible gap ($\le\!0.10$) but the model shows a large one ($\ge\!0.25$), a
segment split a team would ``discover'' and act on that does not exist in real
people. Human shares use the full in-scope table; the model gap carries a
$95\%$ bootstrap CI over respondents.

\textbf{Stability (RQ4): do the conclusions survive irrelevant changes?} We report
the rate of invalid or refused outputs (answers the parser cannot read as a valid
choice), agreement between two decoding runs, and how results move across model
family, model size, and prompt wording.

\subsection{Estimation choices and their robustness}
\label{sec:estimation}
Four choices in the estimation deserve explicit justification, since each is a
place a reader might worry the headline effects are an artifact of the analysis
rather than of the models. We resolve each on the data itself and report the
checks in Section~\ref{sec:rq4} and Appendix~\ref{app:robustness}.

\emph{Confidence level.} We use $95\%$ intervals throughout. The individual-level
deficit (RQ1) and the over-determination effect (RQ3) are large relative to their
uncertainty, so moving from a $90\%$ to a stricter $95\%$ interval leaves every
headline unchanged: on WVS every individual-accuracy margin remains strictly below
zero, on GSS every model remains a statistical tie-or-worse against the baseline,
and the stereotyping index remains significant for the large majority of
question--group pairs (Table~\ref{tab:main}).

\emph{Multiple comparisons.} The stereotyping index is tested over many
(question, axis) pairs, so a per-pair interval does not control the
false-discovery rate. We therefore also apply a Benjamini--Hochberg FDR correction
at $q=0.05$ across all pairs within a model and prompt style. Because the effects
are large, correction barely changes the count of significant pairs (e.g.\ Sonnet
single-answer: $32/32 \to 32/32$ on WVS, $10/12 \to 10/12$ on GSS;
Appendix~\ref{app:robustness}), so ``significant for the majority of pairs'' is not an
artifact of uncorrected multiplicity.

\emph{Respondent clustering.} A minority of respondents answer more than one
question ($7.6\%$ on GSS, $1.8\%$ on WVS), so rows are not fully independent. We
recompute the RQ1 margin with a \emph{cluster} bootstrap that resamples
respondents rather than rows; the $95\%$ intervals are essentially unchanged and
every WVS deficit still excludes zero (Appendix~\ref{app:robustness}).

\emph{Survey weights.} GSS and WVS ship official design/population weights
(\texttt{wtssps}, \texttt{W\_WEIGHT}). We estimate human targets unweighted, and
justify that choice by recomputing, with and without weights on the full
in-scope table, the two human-side quantities the claims rest on: the population
answer distribution (the RQ2 target) and the human $\eta^2$ (the RQ3 baseline
that the stereotyping index subtracts). Weighting moves the human distribution by a
mean Jensen--Shannon divergence of $\le\!0.0003$ and the human $\eta^2$ by at most
$0.007$, both one to three orders of magnitude smaller than the effects we report
(JS gaps of $0.04$--$0.38$; stereotyping indices of $+0.05$ to $+0.7$).
Weighting therefore cannot account for either failure, so we report the simpler
unweighted analysis and document the check in Appendix~\ref{app:robustness}.

%% ======================================================================
\section{Results}
\label{sec:results}

\begin{figure}[!htbp]
\centering
\includegraphics[width=0.49\linewidth]{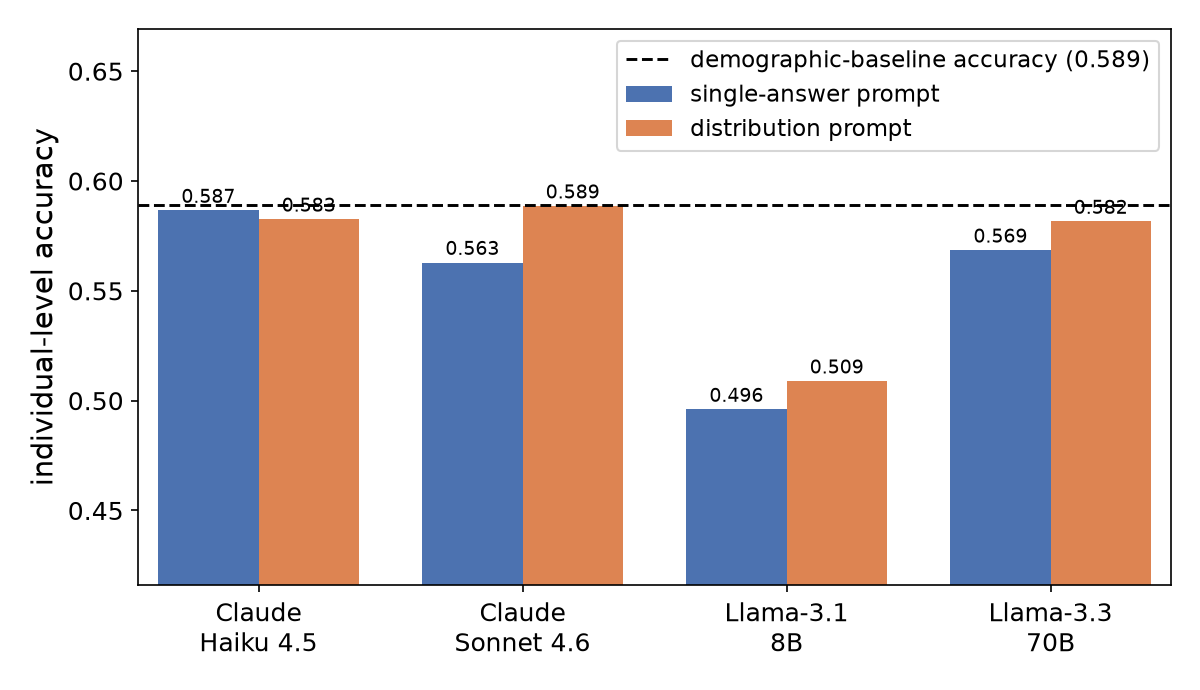}
\hfill
\includegraphics[width=0.49\linewidth]{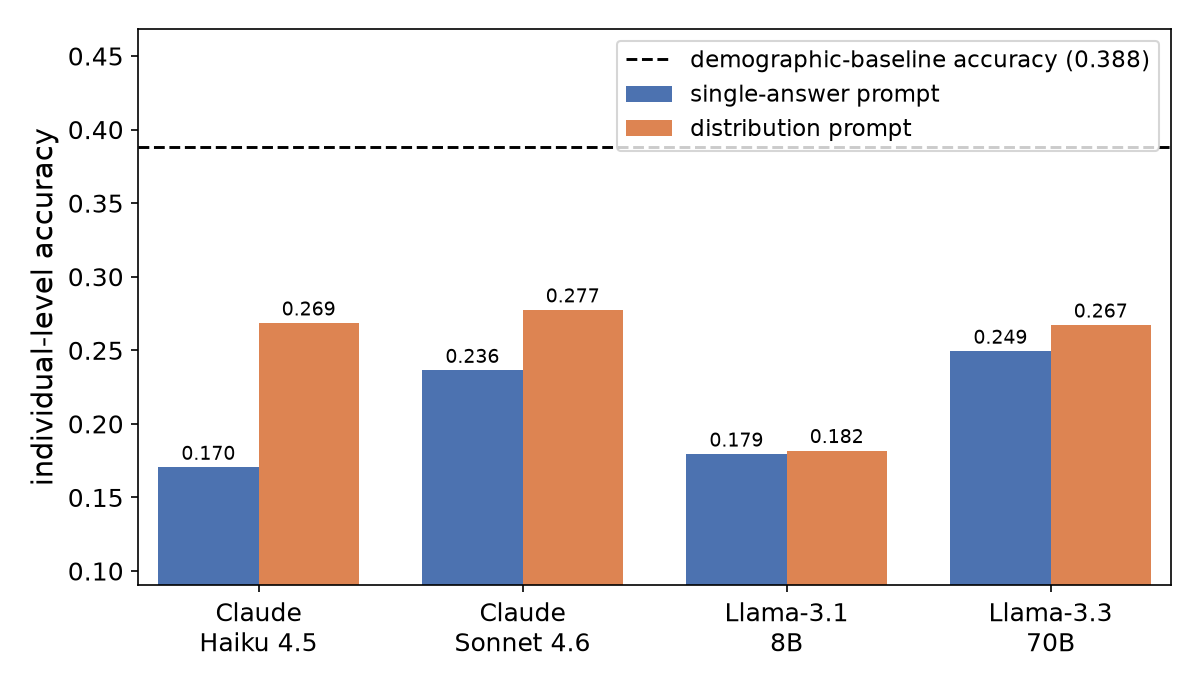}
\caption{Individual accuracy vs.\ the naive demographic baseline (dashed line),
GSS (left) and WVS (right). No model exceeds the baseline on GSS; every model
falls well below it on WVS. The individual-level failure (RQ1) holds in both
domains and is more severe for cross-cultural values.}
\label{fig:indiv}
\end{figure}

\begin{figure}[!htbp]
\centering
\includegraphics[width=0.49\linewidth]{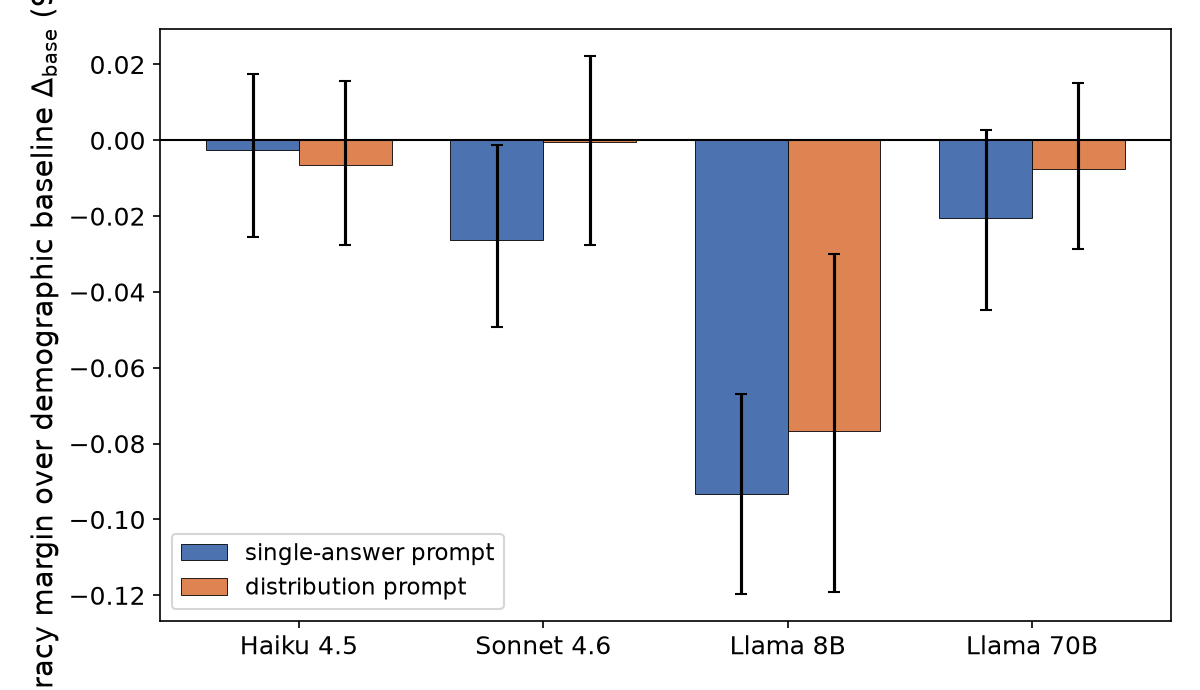}
\hfill
\includegraphics[width=0.49\linewidth]{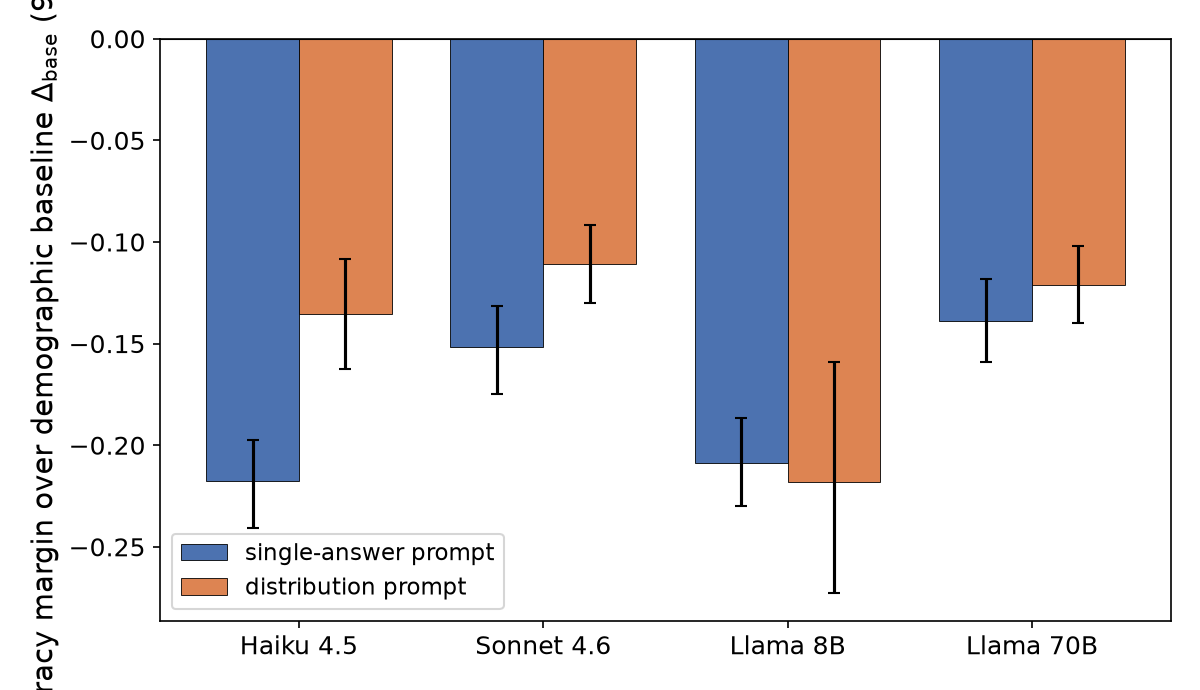}
\caption{Individual-accuracy margin over the demographic baseline
($\Delta_{\text{base}}$) with paired-bootstrap $95\%$ confidence intervals, GSS
(left) and WVS (right). Negative means worse than the baseline. Every interval lies
below zero except the two GSS ties, and the WVS deficits are large and tight.}
\label{fig:deltabase}
\end{figure}

\begin{figure}[!htbp]
\centering
\includegraphics[width=0.49\linewidth]{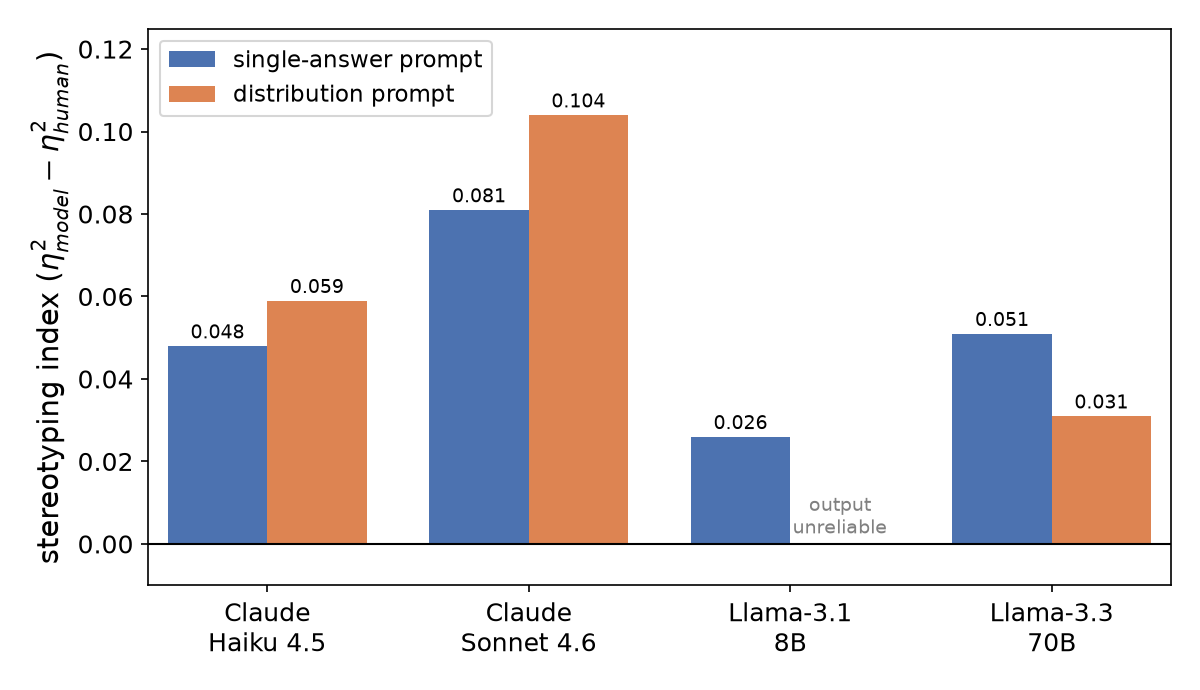}
\hfill
\includegraphics[width=0.49\linewidth]{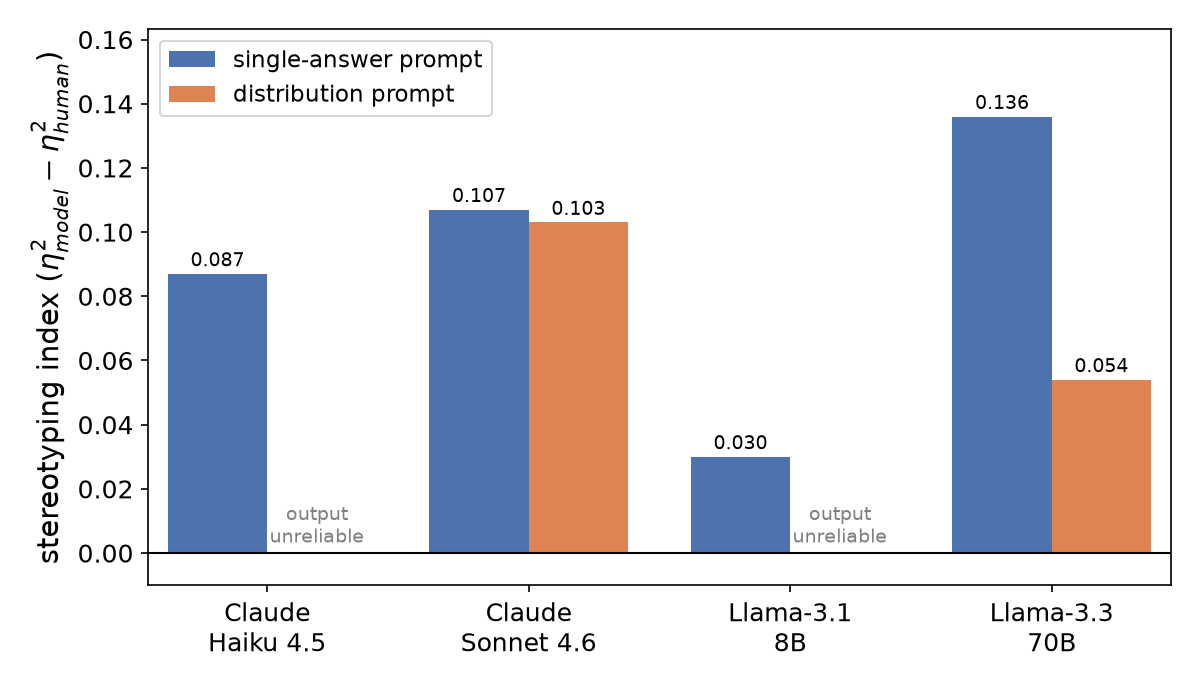}
\caption{Stereotyping index ($\eta^2_{\text{model}}-\eta^2_{\text{human}}$) by
model and prompt style, GSS (left) and WVS (right); higher means the model treats
demographics as more decisive of the answer than they are among real people. All
reliable bars are positive in both domains: every model over-determines
demographics (RQ3), and the larger, more capable models do not reduce the effect
(RQ4). Distribution-prompt bars with high invalid-output rates are omitted and
marked ``output unreliable'' (see Section~\ref{sec:rq4}).}
\label{fig:stereo}
\end{figure}

\begin{figure}[!htbp]
\centering
\includegraphics[width=0.49\linewidth]{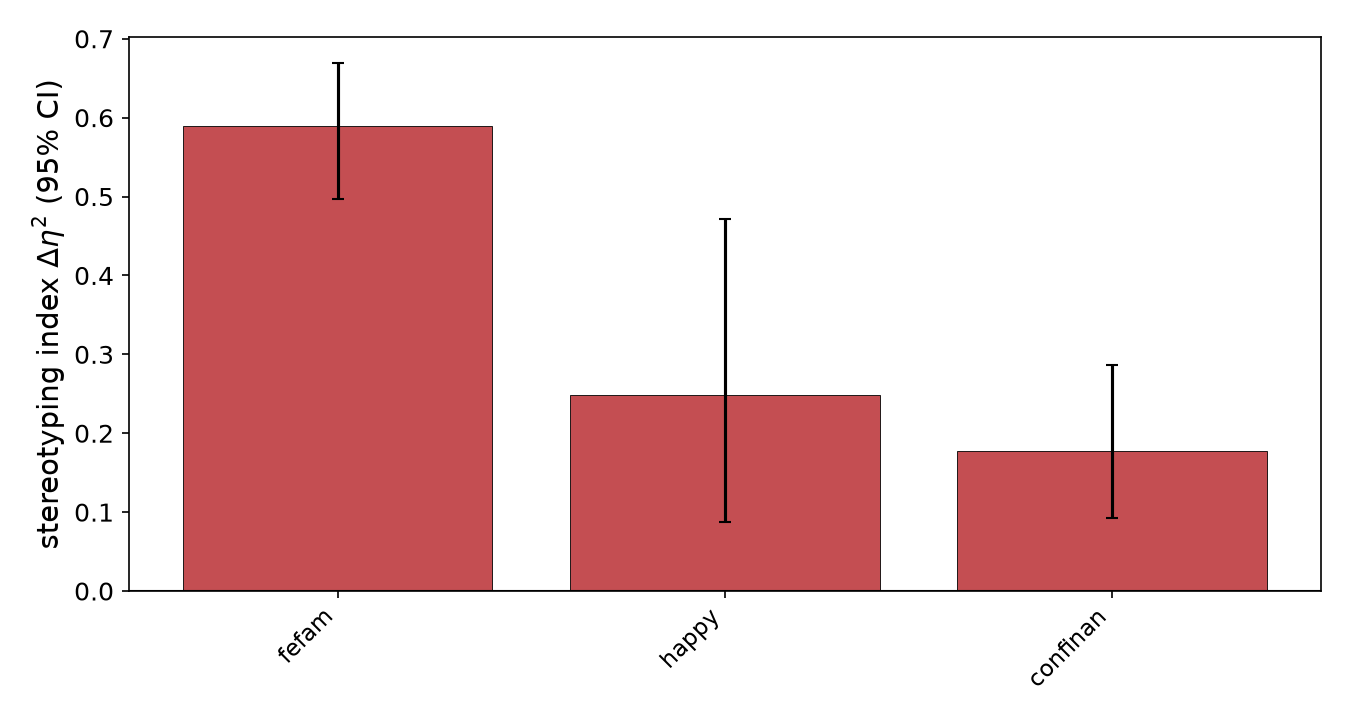}
\hfill
\includegraphics[width=0.49\linewidth]{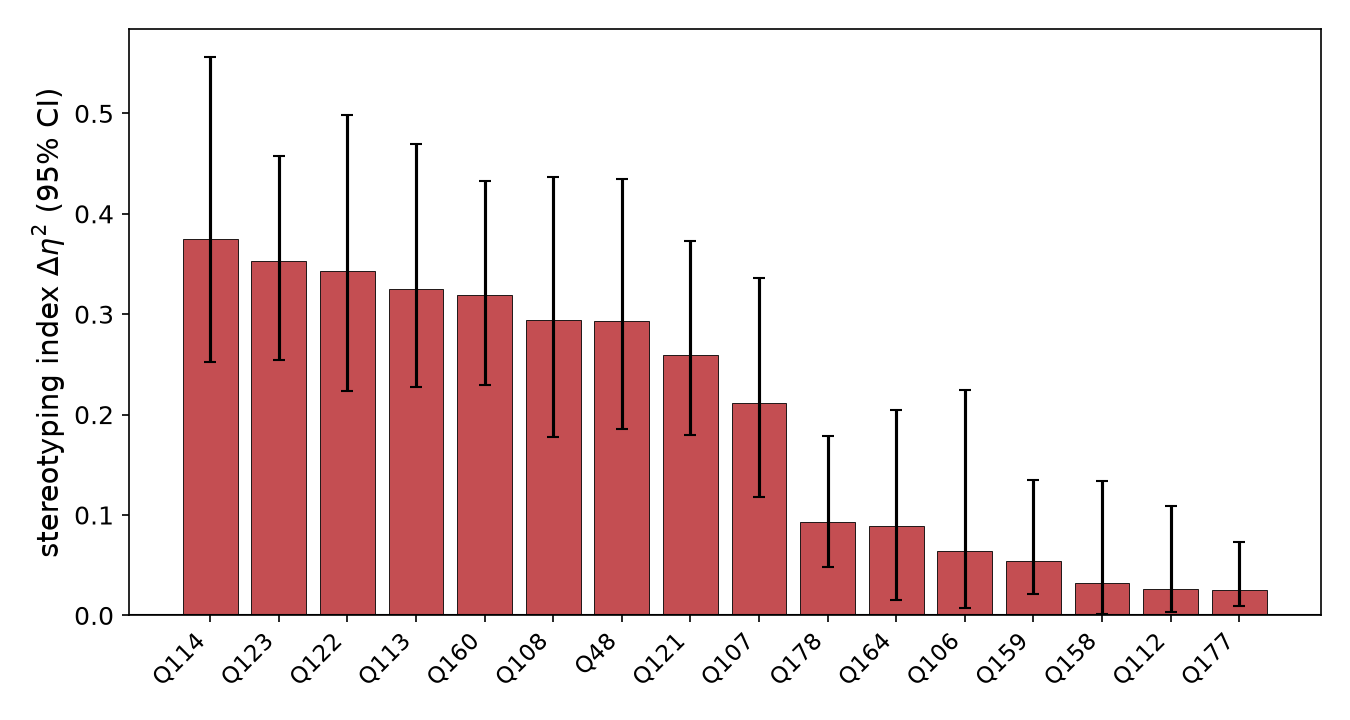}
\caption{Question-level over-determination for the frontier model (Sonnet~4.6,
single-answer prompt) with bootstrap $95\%$ CIs, along the primary grouping axis
(GSS: political views; WVS: country). Over-determination is not a model-average
artifact: it holds question by question, with confidence intervals well above zero
for the strongest items.}
\label{fig:qstereo}
\end{figure}

Table~\ref{tab:main} reports the full cross-model, cross-domain results, and
Table~\ref{tab:baselines} the RQ1 robustness checks (stronger baselines,
distance-aware and proper scoring). The individual-fidelity result is shown in
Figures~\ref{fig:indiv} and~\ref{fig:deltabase} (the latter adding paired-bootstrap
confidence intervals on the accuracy margin over the baseline), and the
over-determination result in Figures~\ref{fig:stereo} and~\ref{fig:qstereo} (the
latter resolving it to the individual-question level).

\begin{table}[!htbp]
\centering
\small
\caption{Cross-domain, cross-model results. $\Delta$base $=$ individual
accuracy minus the demographic-lookup baseline (negative $=$ worse than the
baseline); paired-bootstrap $95\%$ CIs for this margin are shown per model in
Figure~\ref{fig:deltabase}, and the stronger baselines, distance-aware, and
proper-scoring comparisons in Table~\ref{tab:baselines}. JS $=$ mean
Jensen--Shannon divergence (lower is better; population-reweighted values agree, see
Section~\ref{sec:rq2}). $\overline{\Delta\eta^2}$ $=$ median stereotyping index
(positive $=$ over-determination); ``sig'' $=$ number of question--group pairs
whose 95\% CI excludes zero, out of the total (a coding-invariant Cram\'er's~V
replication is in Section~\ref{sec:rq3}). Style~A $=$ single answer, Style~C $=$
distribution.}
\label{tab:main}
\begin{tabular}{llrrrrrl}
\toprule
Domain & Model & Sty & Acc & Base & $\Delta$base & JS & $\overline{\Delta\eta^2}$ (sig) \\
\midrule
GSS & Haiku 4.5   & A & 0.587 & 0.589 & $-0.003$ & 0.090 & $+0.048$ (5/8) \\
GSS & Haiku 4.5   & C & 0.583 & 0.589 & $-0.006$ & 0.011 & $+0.059$ (9/12) \\
GSS & Sonnet 4.6  & A & 0.563 & 0.589 & $-0.026$ & 0.082 & $+0.081$ (10/12) \\
GSS & Sonnet 4.6  & C & 0.589 & 0.589 & $-0.001$ & 0.017 & $+0.104$ (9/12) \\
GSS & Llama 8B    & A & 0.496 & 0.589 & $-0.093$ & 0.056 & $+0.026$ (7/12) \\
GSS & Llama 8B    & C & 0.509 & 0.586 & $-0.077$ & 0.023 & $+0.026$ (3/7)$^\dagger$ \\
GSS & Llama 70B   & A & 0.569 & 0.589 & $-0.021$ & 0.072 & $+0.051$ (6/8) \\
GSS & Llama 70B   & C & 0.582 & 0.589 & $-0.008$ & 0.016 & $+0.031$ (5/8) \\
\midrule
WVS & Haiku 4.5   & A & 0.170 & 0.388 & $-0.218$ & 0.382 & $+0.087$ (30/32) \\
WVS & Haiku 4.5   & C & 0.269 & 0.404 & $-0.135$ & 0.046 & $+0.056$ (21/24)$^\dagger$ \\
WVS & Sonnet 4.6  & A & 0.236 & 0.388 & $-0.152$ & 0.274 & $+0.107$ (31/32) \\
WVS & Sonnet 4.6  & C & 0.277 & 0.388 & $-0.111$ & 0.037 & $+0.103$ (26/28) \\
WVS & Llama 8B    & A & 0.179 & 0.388 & $-0.209$ & 0.232 & $+0.030$ (22/32) \\
WVS & Llama 8B    & C & 0.182 & 0.400 & $-0.218$ & 0.130 & $+0.376$ (1/1)$^\dagger$ \\
WVS & Llama 70B   & A & 0.249 & 0.388 & $-0.139$ & 0.293 & $+0.136$ (31/32) \\
WVS & Llama 70B   & C & 0.267 & 0.388 & $-0.121$ & 0.046 & $+0.054$ (27/30) \\
\bottomrule
\end{tabular}

\vspace{2pt}
{\footnotesize $^\dagger$ Style-C cells with elevated invalid rates
(Section~\ref{sec:rq4}); interpret with caution. Llama-8B Style-C on WVS is
effectively unusable (85\% invalid, $n=440$).}
\end{table}

\subsection{RQ1: LLMs do not beat, and on values fall well below, any non-LLM baseline}
\label{sec:rq1}
On GSS, the individual accuracy of every model is at or below the demographic
lookup baseline ($0.589$). The best margin any model achieves is $-0.001$
(Sonnet, Style~C), a paired-bootstrap tie (CI $[-0.028, 0.022]$); Llama-8B is
$9$ points worse ($0.496$ vs.\ $0.589$, CI $[-0.120, -0.067]$). The lookup is not
a weak straw man: a learned \textbf{logistic} baseline reaches $0.622$ and a random
forest $0.583$ (Table~\ref{tab:baselines}), so the best LLM also trails the strongest
non-LLM predictor. On WVS the gap is far larger and uniformly negative: under the single-answer prompt
every model is \textbf{13.9 to 21.8 accuracy points below} the demographic baseline
(e.g.\ Haiku Style~A: $0.170$ vs.\ $0.388$), and every model under either prompt is
below the learned baselines as well. On WVS every one of these margins has a
$95\%$ CI that lies strictly below zero; on GSS no margin is positive, with the
deficits for Llama-8B (both prompts) and Sonnet's single-answer prompt clearing
$95\%$ significance and the remaining cells statistical ties at or below the
baseline. So for predicting an individual's response, and especially a
cross-cultural value response, a simple demographic predictor beats all four LLMs.

Two robustness checks close the obvious escape routes. First, the deficit is not an
artifact of harsh exact-match scoring: on the ordinal questions the model's
\emph{distance-aware} error is also worse than the baseline's. On GSS the model's
MAE is $0.56$--$0.72$ versus $0.54$ for the baseline; on WVS's long scales it is
$1.83$--$2.27$ versus $1.74$ (Table~\ref{tab:baselines}). The distributional EMD,
which credits a distribution for placing mass near the truth even when its mode is
wrong, tells the same story ($0.76$--$0.83$ vs.\ $0.73$ on GSS; $1.94$--$2.30$ vs.\
$1.67$--$2.02$ on WVS). Giving partial credit for near-misses does not rescue the
model. Second, scoring the \emph{whole} Style~C
distribution with proper scoring rules tells the same story: the model assigns
\emph{lower} probability to the true human answer than the baseline does. On GSS the
model's log-loss is $0.84$--$1.03$ versus $0.81$ for the baseline (Brier $0.52$--$0.62$
vs.\ $0.50$); on WVS $1.76$--$2.26$ versus $1.70$ (Brier $0.80$--$0.89$ vs.\ $0.71$).
This directly answers RQ1: synthetic users provide \emph{no} individual-level
advantage over a demographic baseline under any of these scorings, and on values
they are actively worse.

\begin{table}[!htbp]
\centering
\small
\caption{RQ1 robustness: the individual-level deficit survives stronger baselines,
distance-aware scoring, and proper scoring. \emph{Top:} pooled individual accuracy of
the four non-LLM baselines (the best LLM is shown for reference). \emph{Bottom:}
per-model distance-aware error on ordinal questions and proper scores, each next to
the baseline on the same rows (all lower is better): MAE $=$ mean absolute error of
the single-answer prediction; EMD $=$ earth-mover's distance of the full predicted
distribution (Style~C) from the true answer, in scale steps; log-loss and Brier are
evaluated on the distribution prompt at the true answer. Each cell shows model\,/\,baseline
(m/b) on the same rows. In every cell the model is at best tied and usually worse
than the baseline.}
\label{tab:baselines}
{\small\setlength{\tabcolsep}{5pt}
\begin{tabular}{lrrrrr}
\toprule
& Marginal & Dem.\ lookup & Logistic & Random forest & Best LLM \\
\midrule
GSS accuracy & 0.568 & 0.589 & \textbf{0.622} & 0.583 & 0.589 \\
WVS accuracy & 0.348 & 0.388 & \textbf{0.393} & 0.366 & 0.277 \\
\bottomrule
\end{tabular}
}

\vspace{5pt}
{\small
\begin{tabular}{llcccc}
\toprule
& & MAE & EMD & log-loss & Brier \\
Domain & Model & (m/b) & (m/b) & (m/b) & (m/b) \\
\midrule
GSS & Haiku 4.5  & 0.57 / 0.54 & 0.78 / 0.73 & 0.84 / 0.81 & 0.52 / 0.50 \\
GSS & Sonnet 4.6 & 0.60 / 0.54 & 0.78 / 0.73 & 0.84 / 0.81 & 0.52 / 0.50 \\
GSS & Llama 8B   & 0.68 / 0.54 & 0.83 / 0.74 & 1.03 / 0.80 & 0.62 / 0.50 \\
GSS & Llama 70B  & 0.59 / 0.54 & 0.76 / 0.73 & 0.86 / 0.81 & 0.53 / 0.50 \\
\midrule
WVS & Haiku 4.5  & 1.86 / 1.74 & 1.94 / 1.67 & 1.76 / 1.70 & 0.80 / 0.71 \\
WVS & Sonnet 4.6 & 1.84 / 1.74 & 2.22 / 2.02 & 1.88 / 1.85 & 0.81 / 0.74 \\
WVS & Llama 8B   & 2.05 / 1.74 & 2.30 / 1.87 & 2.26 / 1.85 & 0.89 / 0.73 \\
WVS & Llama 70B  & 1.83 / 1.74 & 2.27 / 2.02 & 1.94 / 1.85 & 0.82 / 0.74 \\
\bottomrule
\end{tabular}
}
\end{table}

\subsection{RQ2: Aggregate distributions are reproduced reasonably, especially under Style C}
\label{sec:rq2}
Reproducing the group's overall answer \emph{spread} is much easier for the models
than guessing individuals, and asking the model directly for a distribution
(Style~C) does this markedly better than asking for a single answer (Style~A).
Recall that the JS divergence runs from $0$ (distributions identical to humans')
upward, so smaller is better. On GSS, Style~C reaches $0.011$--$0.023$ (essentially
the human distribution), versus $0.056$--$0.090$ for Style~A. WVS shows the same
ordering (Style~C $\approx 0.037$--$0.046$ for the models that produce valid
output, versus $0.23$--$0.38$ for Style~A). This confirms, across both domains, the
familiar pattern that a model can mimic ``the average group'' far better than any
individual within it. This aggregate result is not an artifact of the cell-based
sampling: reweighting each cell back to its human population share leaves the JS
divergence essentially unchanged (e.g.\ Haiku GSS Style~C $0.0108 \to 0.0109$;
Sonnet WVS Style~C $0.0371 \to 0.0373$), so the model is being compared against the
true population distribution, not against our sampling design.

\subsection{RQ3: LLMs over-determine demographics (stereotyping), not flatten them}
\label{sec:rq3}
A common worry is that LLMs \emph{flatten} groups, washing out the differences
between them. For identity-linked attitudes we find the opposite. The stereotyping
index is \emph{positive} for essentially every model and prompt style in both
domains, and its confidence interval excludes zero for the large majority of
question--group combinations (Table~\ref{tab:main}): models treat demographics as
\emph{more} decisive of the answer than they are among real people. The effect is
dramatic for politically- and identity-linked questions. On GSS, a person's
political views explain about $10\%$ of the real variation in gender-role
attitudes, but in the models they appear to explain $60$--$69\%$; for confidence
in banks, political views explain only about $1.5\%$ of the real variation yet up
to about $67\%$ in the model, roughly a forty-fold exaggeration. On WVS, taking
country as the grouping, nearly all $32$ question--country combinations are
significantly over-determined for the two Claude models and the larger Llama. The
simulated population is therefore not a blurred copy of the real one; it is a
\emph{caricature}, in which who you are dictates what you think far more tightly
than it does in reality.

This does not depend on how we code the answers. The coding-invariant measure,
Cram\'er's~V, which uses no numeric codes and treats every answer as an unordered
category, tells the same story on \emph{all} questions, including the nominal GSS
items the $\eta^2$ index omits. The median $\Delta V$ is positive for both Claude
models in both domains (GSS Sonnet Style~A $+0.152$, Haiku $+0.108$; WVS Sonnet
Style~A $+0.160$, Llama-70B $+0.166$), and its bootstrap CI excludes zero for a
majority of question--group pairs (e.g.\ $22$ of $40$ for Sonnet on GSS,
$23$ of $32$ on WVS). Over-determination is thus a property of the model's
behavior, not of the numeric coding of the scale.

\subsection{Decision impact: over-determination produces wrong and spurious segment decisions}
\label{sec:decision}
\begin{figure}[!htbp]
\centering
\includegraphics[width=0.49\linewidth]{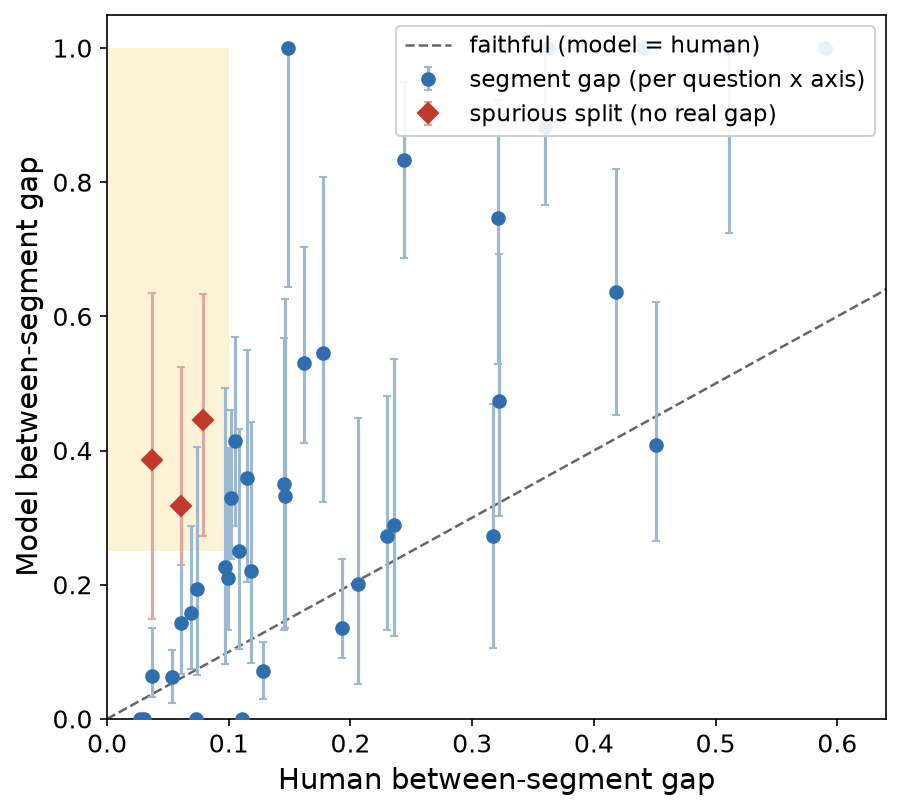}
\hfill
\includegraphics[width=0.49\linewidth]{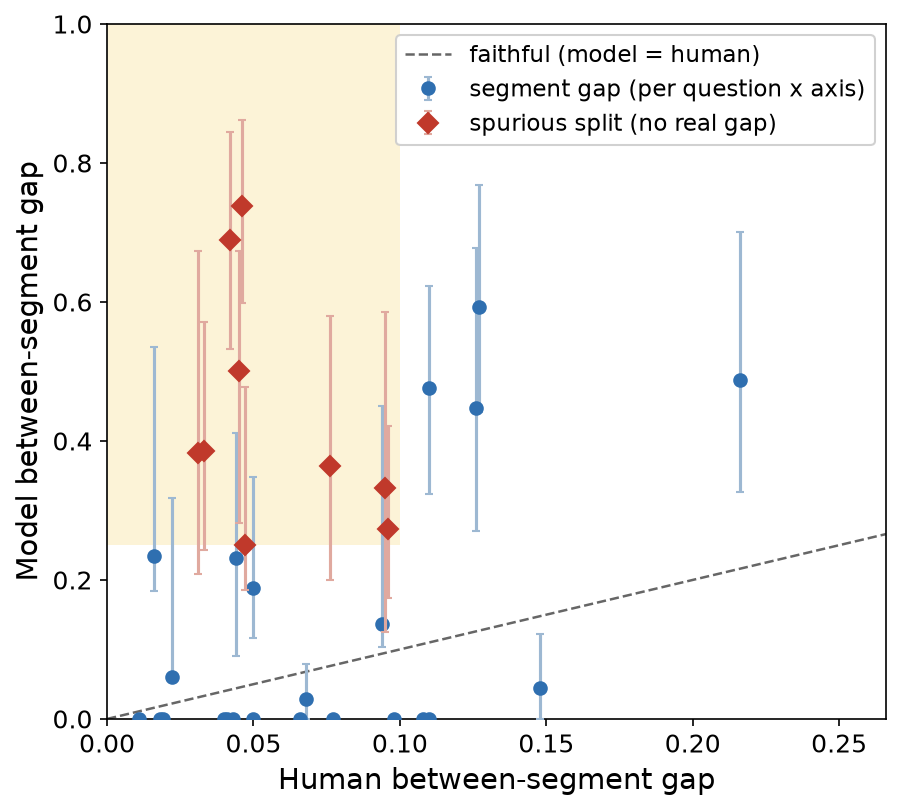}
\caption{Decision impact of over-determination for the frontier model (Sonnet~4.6,
single-answer prompt): the between-segment gap a targeting decision reads off the
model (vertical axis, with $95\%$ bootstrap CI) against the true human gap
(horizontal axis), one point per (question, demographic axis), GSS (left) and WVS
(right). The dashed line is faithful reading (model $=$ human); points above it are
gaps the model exaggerates. Shaded: the \emph{spurious-split} zone (human gap
$\le\!0.10$ but model gap $\ge\!0.25$), the segment differences a team would act on that
do not exist in real people. Almost every point lies above the faithful line.}
\label{fig:decision}
\end{figure}

Over-determination is not just a statistical curiosity; it changes what a team
would decide. Reading segment structure off the synthetic users rather than off
real humans distorts the targeting decision in three concrete ways
(Figure~\ref{fig:decision}). First, the model \textbf{inflates the between-segment
gap} a decision keys on: for the frontier model (Sonnet, single-answer) the median
gap is $2.3\times$ the true human gap on GSS and $2.5\times$ on WVS, and every
model inflates it (medians $1.3$--$4.1\times$ on GSS, $2.0$--$4.7\times$ on WVS). A
team sizing a segment difference off synthetic users over-estimates it roughly
two to fourfold. Second, the model often \textbf{targets the wrong segment}: its
$\arg\max$ segment differs from the true human $\arg\max$ for $50\%$ of GSS
(question, axis) pairs and $72\%$ of WVS pairs (Sonnet single-answer), so a team
choosing which segment to prioritize would, more often than not on values, pick a
segment that is not actually the most extreme. Third, and most damaging, the model
\textbf{manufactures segment splits that do not exist}: on WVS, $28\%$ of pairs
(Sonnet single-answer; up to $41\%$ for Haiku and Llama-70B) fall in the
spurious-split regime: humans differ across segments by $\le\!10$ points while the
model differs by $\ge\!25$. Concrete cases make the failure vivid. Among U.S.\
respondents, confidence in banks barely varies with political leaning (a
``great deal'' of confidence ranges only $13$--$24\%$ across the seven political
categories), yet the model treats politics as roughly $67\%$ determinative of the
answer; a team reading the simulation would conclude, wrongly, that trust in banks
is sharply polarized. On WVS, attitudes to government responsibility (Q108) differ
across education levels by under $5$ points among real people but by $74$ points in
the model ($16\times$; CI $[0.60, 0.86]$), a segment split with no basis in the
human data. Because these errors are \emph{structured}, always in the direction of
exaggerated group difference, they do not average out across a portfolio of
decisions; they systematically push a team toward over-segmenting a population and
over-targeting the segment the model has caricatured.

\subsection{RQ4: Capability does not fix it; outputs are unstable to format and prompt surface}
\label{sec:rq4}
Three stability results stand out. First, \textbf{capability does not remedy either
failure}. Within the closed family, the frontier model (Sonnet~4.6) stereotypes
\emph{more} than the smaller Haiku~4.5 (median $\Delta\eta^2$ $+0.104$ vs.\
$+0.059$ on GSS Style~C; $+0.103$ vs.\ $+0.056$ on WVS Style~C), and does not beat
the baseline on individuals. Within the open family, Llama-70B has the highest
stereotyping index on WVS ($+0.136$). Larger, more capable models are therefore
not safer synthetic users on these axes. Second, \textbf{format stability is a
real failure mode for smaller models}. The distribution prompt (Style~C) over
long numeric scales produced high invalid-output rates: Llama-8B reached 85\%
invalid on WVS (rendering its Style-C numbers unusable, $n=440$) and 57\% on GSS;
Haiku reached 40\% on WVS. Sonnet and Llama-70B remained at 0\% invalid. This is
itself a practical result: a researcher relying on distribution prompts from a
small model may silently lose most of their sample.

The accuracy numbers elsewhere in the paper are \emph{conditional on valid output},
which flatters the high-invalid cells because invalid responses may not be missing
at random. We therefore report both bounds. Where invalid rates are near zero (all
Claude and Llama-70B cells), the two coincide, so those results are unaffected.
Where they are high, treating each invalid output as a failed prediction, the
appropriate accounting when a synthetic user must return a usable answer, widens
the deficit sharply: Llama-8B's Style-C accuracy falls from $0.51$ to $0.22$ on GSS
and from $0.18$ to $0.03$ on WVS, and Haiku's WVS Style-C from $0.27$ to $0.16$.
Because the non-LLM baselines never emit invalid output, penalizing invalids can
only widen the LLM's gap to the baseline, never close it; the individual-level
conclusion is robust to either accounting.

Third, \textbf{predictions are unstable to prompt-surface choices that should be
irrelevant}. Prior work shows that survey-style LLM answers are sensitive to option
ordering and labeling~\cite{dominguezolmedo2024questioning} and that prompt
architecture and persona formulation can themselves induce
artifacts~\cite{brucks2025prompt,lutz2025prompt}; we quantify this for our setting.
On the two clean closed models we measured the \emph{prediction-flip rate} (the
fraction of single-answer predictions that change) under two perturbations that
leave the actual question and answer set unchanged: reversing the display order of
the answer options, and reframing the identical demographics as a natural-language
persona sentence rather than a bulleted profile. We benchmark these against a decoding-noise floor: the flip rate
between two decoding seeds of the \emph{same} prompt. On GSS, merely reversing
option order flips $9.8\%$ (Sonnet) to $10.8\%$ (Haiku) of predictions, and the
persona reframing flips a further $\sim\!9\%$, both roughly $4$--$10\times$ the
seed-noise floor ($1.1\%$ and $2.6\%$ respectively). An option-order spot-check on
WVS shows the same effect more strongly ($14.3\%$ flips for Sonnet, $22.6\%$ for
Haiku). Option order also moves aggregate fidelity (e.g.\ Haiku GSS accuracy
$0.587\!\rightarrow\!0.606$), so a study's conclusions can depend on an arbitrary
presentation choice. Together with the capability and format results, this means
synthetic-user validity is not only a model property but a prompt/interface
property.

\subsection{RQ5: Both failures transfer across domains}
\label{sec:rq5}
The two headline failures (no individual-level advantage over the demographic
baseline, and positive stereotyping) appear in \emph{both} the U.S.\ social-attitude
domain and the 63-country values domain, for \emph{all four} models and \emph{both}
families. The individual-level failure is in fact \emph{stronger} in the
cross-cultural domain (an 11--22 point deficit vs.\ a rough tie). This cross-domain
consistency is the basis for treating these as properties of current LLM
synthetic-user methods rather than artifacts of a single survey, model, or prompt.

%% ======================================================================
\section{Discussion}
\label{sec:discussion}

%\paragraph{What these results mean for using synthetic users}
The two failures discussed in this paper stem from the same underlying cause: current LLMs, prompted with demographics,
appear to answer by mapping a profile to a stereotyped ``typical'' response for
that profile. This is consistent with evidence that language models encode
social associations~\cite{caliskan2017semantics,bender2021dangers} and that persona
conditioning steers models toward stereotyped abstractions rather than nuanced
individuals~\cite{deshpande2023toxicity}. That mapping is too coarse to beat a
demographic lookup at the individual level, and it is worse than the lookup when the
answer space is large, as in the 10-point WVS scales. Yet it is also too tight: it
treats demographics as more predictive of attitudes than they truly are. The
practical corollary is that a synthetic-user study can look successful on an
aggregate-distribution metric, even one improved by distribution-specialized
training~\cite{cao2025}, while being invalid for any individual- or subgroup-level
use, and while systematically exaggerating the demographic structure of opinion.
Because the aggregate metric is the one most often reported, this is a live risk for
the field.

\paragraph{Implications for decision support}
These two failures map directly onto how synthetic users are used to support
decisions. A team that only checks aggregate fidelity, the common practice, can
be misled in two ways at once. Any decision keyed to \emph{individuals or narrow
segments} (personalizing an offer, sizing a niche, or predicting how a specific
demographic will react) rests on individual-level fidelity the models do not have;
a demographic lookup would serve at least as well and often better, at a fraction of
the cost. And any decision that reads the \emph{structure} of opinion off the
simulation (``this attitude splits sharply along political lines,'' ``this segment
is monolithic'') inherits the models' over-determination, which manufactures
group differences that are far larger than they are in real people. Our
decision-impact analysis (Section~\ref{sec:decision}) puts numbers on this second
failure: on the segment-targeting task the models inflate between-segment gaps two
to fourfold, would send a team to the wrong segment in half of GSS and most WVS
cases, and manufacture a nonexistent segment split in up to $41\%$ of value
questions. The danger is not random noise, which averages out, but this
\emph{structured} error, which propagates into the decision. The framework we
propose is meant to catch exactly
this before the evidence is trusted: it turns ``did the aggregate look right?'' into
a set of checks a team can run against held-out human data to decide, for their
specific question, whether synthetic-user evidence is safe.

\paragraph{A validation protocol}
From these results we distill a checklist for researchers considering LLM
synthetic users: (1) report individual-level and aggregate-level fidelity
\emph{separately}; (2) always benchmark individual fidelity against non-LLM
baselines computed on real held-out data (at least a demographic lookup, ideally a
learned demographic model) using distance-aware or proper-scoring metrics on
ordinal scales; an LLM that does not beat them adds no individual-level information;
(3) report subgroup determinism (e.g.\ $\Delta\eta^2$, or a coding-invariant
association measure), not just group means, to detect stereotyping; (4) before
acting on any segment-level read, check the decision-impact quantities (gap
inflation, wrong-target rate, and spurious-split rate) against held-out human
data, since a distortion invisible in an aggregate metric can still flip a
targeting decision; (5) report invalid/refusal rates, especially for distribution
prompts and smaller models; (6) do not assume a larger or more capable model is a
safer synthetic user; and (7) restrict validity claims to the population, domain,
and level of analysis actually tested.

\paragraph{Limitations}
Our claims are scoped to the data we used: GSS respondents 2016--2024 and WVS
Wave~7 across 63 countries, on the specific questions listed, with two prompt
formats and two seeds, and to demographic prompting under survey-simulation
protocols, not to every elicitation strategy. The LLM-evaluated sample is about
$100$ respondents per question, but the individual-fidelity claims pool all rows in
a domain ($n\approx2{,}000$ on GSS, $n\approx2{,}900$ on WVS) and every headline
comparison carries a paired-bootstrap $95\%$ interval, so the conclusions rest on
those intervals rather than on any single question's cell. We do not claim these
numbers characterize every model, prompt, or elicitation strategy; richer persona
construction~\cite{li2025persona}, few-shot conditioning, or fine-tuning
could shift the individual-fidelity and stereotyping numbers, and testing that is
exactly the value of a fixed, reusable protocol. We estimate human targets without
survey weights; because applying the official GSS and WVS weights moves those
targets by amounts one to three orders of magnitude smaller than the effects we
report (Appendix~\ref{app:robustness}), this choice does not affect any claim. Older public
surveys such as these may appear in model pretraining data; we mitigate this by
scoping GSS to recent waves and by noting that memorization would, if anything,
\emph{help} aggregate fidelity while leaving the individual-level and
over-determination failures intact. Finally, survey answers are stated opinions,
not behavior; extending the protocol to behavioral tasks and to interactive
multi-agent simulations~\cite{park2023generative,mou2026from} is future work.

%% ======================================================================
\section{Conclusion}
\label{sec:conclusion}

Across two independent domains, four models, and two model families, and under the
demographic-prompting protocols we test, LLM synthetic users show two robust
failures: they do not beat, and on cross-cultural values fall well below, non-LLM
baselines at the individual level, and they systematically over-determine
demographics, treating identity as far more predictive of attitudes than it is
among real people. The individual-level deficit survives distance-aware and proper
scoring, and the over-determination survives a coding-invariant association measure;
model capability does not fix either failure.

The contribution, however, is not merely the negative result that these systems
fail. It is a reusable evaluation framework for intelligent synthetic-user
systems: the cross-domain benchmark, the suite of non-LLM baselines scored with
distance-aware and proper-scoring metrics, the stereotyping index and its
coding-invariant companion, and the validation protocol that ties them
together, designed to be run before synthetic-user evidence is deployed in a
decision-support workflow. Rather than trusting an aggregate similarity number, a
team can use this framework to determine, for their specific question, population,
and level of analysis, whether an LLM synthetic user is fit for purpose or whether a
trivial demographic predictor would serve at least as well. As LLMs are increasingly
wired into product, policy, and market decisions, that up-front check is what turns a
cautionary finding into an actionable safeguard.

%% ======================================================================
\appendix
\sloppy

\section{Model call settings}
\label{app:models}
All four models (Claude Haiku~4.5, Claude Sonnet~4.6, Llama-3.1-8B-Instruct, and
Llama-3.3-70B-Instruct) are called through a single common chat-completion
interface. Decoding uses temperature $1.0$ (the
model's default sampling; where a model rejects an explicit temperature field it is
omitted and the provider default applies) and a token cap of $100$. Each
(model, prompt format, respondent, question) cell is generated under two independent
runs (``seeds'' $0$ and $1$); these differ only in sampling randomness and are used
to estimate the run-to-run noise floor in RQ4.

\section{Exact questions and answer scales}
\label{app:questions}
Tables~\ref{tab:gssq} and~\ref{tab:wvsq} give every question verbatim, with its
answer options and integer coding, whether it is treated as ordinal or nominal, and
which metrics it enters. GSS wordings and value labels are the standard GSS codebook
text; WVS wordings and ordinal scales follow WorldValuesBench for comparability. An
item is ordinal when a mean answer is meaningful; only ordinal items enter the
$\eta^2$ stereotyping index and the distance-aware metrics (MAE/EMD), while all items
enter accuracy, JS divergence, and the coding-invariant Cram\'er's~V ($V$). GSS prompt
demographics are age, sex, race, highest degree, region, political views, and party
identification (subgroup axes: sex, race, degree, region, political views); WVS prompt
demographics are age group, sex, education level, settlement type, and country
(subgroup axes: country, education, age group, sex).

{\footnotesize\setlength{\tabcolsep}{4pt}
\begin{longtable}{@{}p{0.10\linewidth}>{\RaggedRight\arraybackslash}p{0.30\linewidth}>{\RaggedRight\arraybackslash}p{0.23\linewidth}>{\RaggedRight\arraybackslash}p{0.09\linewidth}>{\RaggedRight\arraybackslash}p{0.14\linewidth}@{}}
\caption{GSS questions: wording, answer options (code=label), type, and metrics
entered. ``all'' $=$ acc, JS, MAE/EMD, log-loss/Brier, $\eta^2$, $V$.}\label{tab:gssq}\\
\toprule
Var & Question wording & Options (code=label) & Type & Metrics \\
\midrule
\endfirsthead
\multicolumn{5}{@{}l}{\footnotesize\emph{Table~\ref{tab:gssq} continued}}\\
\toprule
Var & Question wording & Options (code=label) & Type & Metrics \\
\midrule
\endhead
\bottomrule
\endfoot
\texttt{happy} & Taken all together, how would you say things are these days -- would you say that you are very happy, pretty happy, or not too happy? & 1=very happy; 2=pretty happy; 3=not too happy & ordinal & all \\
\texttt{trust} & Generally speaking, would you say that most people can be trusted or that you can't be too careful in dealing with people? & 1=most people can be trusted; 2=can't be too careful; 3=depends & nominal & acc, JS, $V$ \\
\texttt{fair} & Do you think most people would try to take advantage of you if they got a chance, or would they try to be fair? & 1=would take advantage of you; 2=would try to be fair; 3=depends & nominal & acc, JS, $V$ \\
\texttt{helpful} & Would you say that most of the time people try to be helpful, or that they are mostly just looking out for themselves? & 1=try to be helpful; 2=looking out for themselves; 3=depends & nominal & acc, JS, $V$ \\
\texttt{cappun} & Do you favor or oppose the death penalty for persons convicted of murder? & 1=favor; 2=oppose & nominal & acc, JS, $V$ \\
\texttt{grass} & Do you think the use of marijuana should be made legal or not? & 1=should be legal; 2=should not be legal & nominal & acc, JS, $V$ \\
\texttt{abany} & Please tell me whether or not you think it should be possible for a pregnant woman to obtain a legal abortion if the woman wants it for any reason. & 1=yes; 2=no & nominal & acc, JS, $V$ \\
\texttt{gunlaw} & Would you favor or oppose a law which would require a person to obtain a police permit before he or she could buy a gun? & 1=favor; 2=oppose & nominal & acc, JS, $V$ \\
\texttt{fefam} & Do you agree or disagree with this statement: It is much better for everyone involved if the man is the achiever outside the home and the woman takes care of the home and family? & 1=strongly agree; 2=agree; 3=disagree; 4=strongly disagree & ordinal & all \\
\texttt{confinan} & How much confidence do you have in the people running banks and financial institutions -- a great deal, only some, or hardly any? & 1=a great deal; 2=only some; 3=hardly any & ordinal & all \\
\end{longtable}
}

{\footnotesize\setlength{\tabcolsep}{4pt}
\begin{longtable}{@{}p{0.06\linewidth}>{\RaggedRight\arraybackslash}p{0.60\linewidth}p{0.08\linewidth}p{0.05\linewidth}p{0.08\linewidth}@{}}
\caption{WVS questions (WorldValuesBench probe set): wording, scale, type, and metrics
entered. All are ordinal integer scales.}\label{tab:wvsq}\\
\toprule
Var & Question wording & Scale & Type & Metrics \\
\midrule
\endfirsthead
\multicolumn{5}{@{}l}{\footnotesize\emph{Table~\ref{tab:wvsq} continued}}\\
\toprule
Var & Question wording & Scale & Type & Metrics \\
\midrule
\endhead
\bottomrule
\endfoot
\texttt{Q48} & On a scale of 1 to 10, 1 meaning `None at all' and 10 meaning `A great deal', how much freedom of choice and control over your life do you feel you have? & 1--10 & ord. & all \\
\texttt{Q106} & On a scale of 1 to 10, 1 meaning `Incomes should be made more equal' and 10 meaning `There should be greater incentives for individual effort', where would you place your view? & 1--10 & ord. & all \\
\texttt{Q107} & On a scale of 1 to 10, 1 meaning `Private ownership of business should be increased' and 10 meaning `Government ownership of business should be increased', where would you place your view? & 1--10 & ord. & all \\
\texttt{Q108} & On a scale of 1 to 10, 1 meaning `The government should take more responsibility to ensure that everyone is provided for' and 10 meaning `People should take more responsibility to provide for themselves', where would you place your view? & 1--10 & ord. & all \\
\texttt{Q112} & On a scale of 1 to 10, 1 meaning `No corruption at all' and 10 meaning `Abundant corruption', how much corruption do you think there is in your country? & 1--10 & ord. & all \\
\texttt{Q113} & On a scale of 1 to 4, 1 meaning `None of them' and 4 meaning `All of them', how many state authorities do you think are involved in corruption? & 1--4 & ord. & all \\
\texttt{Q114} & On a scale of 1 to 4, 1 meaning `None of them' and 4 meaning `All of them', how many business executives do you think are involved in corruption? & 1--4 & ord. & all \\
\texttt{Q121} & On a scale of 1 to 5, 1 meaning `Very bad' and 5 meaning `Very good', what impact do you think immigrants have on the development of your country? & 1--5 & ord. & all \\
\texttt{Q122} & On a scale of 0 to 2, 0 meaning `Disagree' and 2 meaning `Agree', do you agree that immigration fills useful jobs in the labour market? & 0--2 & ord. & all \\
\texttt{Q123} & On a scale of 0 to 2, 0 meaning `Disagree' and 2 meaning `Agree', do you agree that immigration increases the crime rate? & 0--2 & ord. & all \\
\texttt{Q158} & On a scale of 1 to 10, 1 meaning `Completely disagree' and 10 meaning `Completely agree', how much do you agree that science and technology make our lives healthier, easier, and more comfortable? & 1--10 & ord. & all \\
\texttt{Q159} & On a scale of 1 to 10, 1 meaning `Completely disagree' and 10 meaning `Completely agree', how much do you agree that because of science and technology there will be more opportunities for the next generation? & 1--10 & ord. & all \\
\texttt{Q160} & On a scale of 1 to 10, 1 meaning `Completely disagree' and 10 meaning `Completely agree', how much do you agree that we depend too much on science and not enough on faith? & 1--10 & ord. & all \\
\texttt{Q164} & On a scale of 1 to 10, 1 meaning `Not at all important' and 10 meaning `Very important', how important is God in your life? & 1--10 & ord. & all \\
\texttt{Q177} & On a scale of 1 to 10, 1 meaning `Never justifiable' and 10 meaning `Always justifiable', how justifiable do you think it is to claim government benefits to which you are not entitled? & 1--10 & ord. & all \\
\texttt{Q178} & On a scale of 1 to 10, 1 meaning `Never justifiable' and 10 meaning `Always justifiable', how justifiable do you think it is to avoid paying a fare on public transport? & 1--10 & ord. & all \\
\end{longtable}
}
Here ``all'' denotes accuracy, JS, MAE/EMD, log-loss/Brier, $\eta^2$, and $V$, since
every WVS item is ordinal.

\section{Prompt templates and demographic formatting}
\label{app:prompts}
Demographics are rendered as a bulleted profile, one line per attribute, using
human-readable value labels (e.g.\ ``\texttt{- Political views (liberal--conservative):
conservative}''); for the persona perturbation (RQ4) the same attributes are rewritten
as a single natural-language sentence (``\emph{a person with \dots}''). The two prompt
formats share this demographic block and differ only in the response instruction:

\emph{Single-answer prompt (Style~A).}
\begin{quote}\footnotesize\ttfamily
You are simulating a single \{respondent noun\} with the following demographic
profile:\\[2pt]
\{demographics\}\\[2pt]
Based only on this profile, predict how this specific person would answer the
following survey question.\\[2pt]
Question:\\ \{question text\}\\[2pt]
Answer options:\\ \{options\}\\[2pt]
Respond with only the single \{option letter $\vert$ number\} (e.g. A). No explanation.
\end{quote}

\emph{Distribution prompt (Style~C).}
\begin{quote}\footnotesize\ttfamily
Consider a \{respondent noun\} with the following demographic profile:\\[2pt]
\{demographics\}\\[2pt]
Estimate the probability that this respondent would choose each answer option for the
question below. Probabilities must sum to 1.\\[2pt]
Question:\\ \{question text\}\\[2pt]
Answer options:\\ \{options\}\\[2pt]
Return only valid JSON mapping each answer \{key\} to a probability, using exactly
these keys: \{keys\}. No other text.
\end{quote}

Answer options use short letters (A, B, \dots) when the options have text labels
(GSS), and the scale number itself when options are numeric points on an ordinal
scale (WVS), which avoids the eight-letter limit on the longest scales. In every case
the prompt carries an explicit key$\rightarrow$answer-code mapping that the parser
uses to recover the underlying code.

\section{Parser rules and invalid-output handling}
\label{app:parser}
Outputs are parsed deterministically. For the single-answer prompt with letter keys,
the parser takes the first standalone capital letter that is a valid option key; with
numeric keys it takes the first integer token that is a valid key, matching multi-digit
tokens so that ``10'' is not read as ``1''. For the distribution prompt, the parser
extracts the first balanced JSON object, reads a probability for each declared key
(absent keys default to $0$), and normalizes to sum to one; any response that is empty,
contains no valid key, fails JSON parsing, or yields a non-positive total is flagged
\emph{invalid/refused}. Invalid outputs are excluded from the accuracy and
distribution metrics (a conditional-on-valid accounting) and counted in the
invalid-rate stability metric (RQ4). Because invalids may not be missing at random,
Section~\ref{sec:rq4} also reports the opposite bound (each invalid counted as a
failed prediction), and the two accountings coincide wherever the invalid rate is
near zero. Prose-wrapped answers are tolerated as long as a valid key is recoverable.

\section{Baseline construction and back-off}
\label{app:baselines}
All baselines are fit on the 50\% ``fit'' fold (respondents with an even hash of their
id) and evaluated on the evaluation-sample rows, which are drawn from the ``eval'' fold, so no
respondent is used for both. The \emph{question-marginal} baseline predicts the single
most frequent fit-fold answer for the question. The \emph{demographic-lookup} baseline
estimates a conditional answer distribution for each demographic cell and predicts its
mode, backing off when a cell has fewer than $20$ fit-fold respondents: for GSS the
back-off is \{degree, race, sex, political views\} $\rightarrow$ \{degree, race\}
$\rightarrow$ question marginal; for WVS it is \{country, education, age group\}
$\rightarrow$ \{country\} $\rightarrow$ question marginal, and we report the
\{country\} level separately as a country-only baseline. The \emph{logistic} baseline
is a multinomial logistic regression on the one-hot-encoded prompt demographics, fit per
question ($\le\!2000$ iterations); the \emph{random-forest} baseline uses $300$ trees on
the same features. The lookup and logistic baselines also emit a full predicted
distribution, used as the reference for the log-loss and Brier comparisons.

\section{Estimation robustness}
\label{app:robustness}
This appendix records the four estimation checks summarized in
Section~\ref{sec:estimation}; all are computed on the same cached outputs as the
main results, with no additional model calls.

\emph{Multiple comparisons (FDR).} For each (question, axis) pair entering the
$\eta^2$ stereotyping index we compute a one-sided bootstrap $p$-value for
$H_0\!:\Delta\eta^2\le 0$ and apply a Benjamini--Hochberg correction at $q=0.05$
across all pairs within a model and prompt style. The number of significant,
positively-signed pairs is nearly identical before and after correction, because
the effects are large: for the single-answer prompt, Sonnet is $32/32$ (WVS) and
$10/12$ (GSS) both before and after correction; Haiku $31/32 \to 31/32$ (WVS);
Llama-70B $31/32 \to 31/32$ (WVS) and $7/8 \to 7/8$ (GSS). Over-determination is
therefore not an artifact of uncorrected multiple testing.

\emph{Respondent-clustered bootstrap.} The individual-fidelity sample contains a
small share of respondents who answer more than one question ($7.6\%$ on GSS,
$1.8\%$ on WVS). Recomputing the RQ1 accuracy margin over the baseline with a
cluster bootstrap (resampling respondents, then taking all of their rows) leaves
the $95\%$ intervals essentially unchanged from the row-level intervals: on WVS
every margin still excludes zero (e.g.\ Sonnet single-answer $-0.152$,
$[-0.180,-0.121]$; Haiku $-0.218$, $[-0.248,-0.187]$), and on GSS the Llama-8B
deficits remain significant while the Claude cells remain ties, exactly as in the
row-level analysis. Within-respondent correlation does not drive the result.

\emph{Survey-weight robustness.} GSS and WVS provide official weights
(\texttt{wtssps}, coverage $100\%$; \texttt{W\_WEIGHT}, coverage $100\%$ on the
in-scope pool). Applying them to the two human-side quantities the claims depend on
changes them negligibly: the human population answer distribution moves by a mean
Jensen--Shannon divergence of $0.0001$ (GSS) and $<\!0.0001$ (WVS), with a maximum
over questions of $0.0003$; the human $\eta^2$ that the stereotyping index
subtracts moves by a mean absolute $0.0016$ (GSS) and $0.0012$ (WVS), with a
maximum of $0.0065$. These are one to three orders of magnitude below the effects
we report, so the unweighted analysis is adequate for every claim in the paper.

\emph{Sample adequacy.} Individual-fidelity margins pool all evaluation rows per
domain ($n=1{,}986$ scored (respondent, question) pairs on GSS, $n=2{,}916$ on WVS
for the clean cells) rather than the $\approx\!100$ respondents of any single
question, and every headline comparison carries a paired-bootstrap $95\%$ interval
computed on those rows; the intervals, not a nominal per-question count, are what
license the claims.

%% ======================================================================
\bibliographystyle{unsrtnat}
\bibliography{refs}

@inproceedings{worldvaluesbench2024,
  title={Worldvaluesbench: A large-scale benchmark dataset for multi-cultural value awareness of language models},
  author={Zhao, Wenlong and Mondal, Debanjan and Tandon, Niket and Dillion, Danica and Gray, Kurt and Gu, Yuling},
  booktitle={Proceedings of the 2024 Joint International Conference on Computational Linguistics, Language Resources and Evaluation (LREC-COLING 2024)},
  pages={17696--17706},
  year={2024}
}

@article{argyle2023,
  title={Out of one, many: Using language models to simulate human samples},
  author={Argyle, Lisa P and Busby, Ethan C and Fulda, Nancy and Gubler, Joshua R and Rytting, Christopher and Wingate, David},
  journal={Political Analysis},
  volume={31},
  number={3},
  pages={337--351},
  year={2023},
  publisher={Cambridge University Press}
}

@article{bisbee2024,
  title={Synthetic replacements for human survey data? The perils of large language models},
  author={Bisbee, James and Clinton, Joshua D and Dorff, Cassy and Kenkel, Brenton and Larson, Jennifer M},
  journal={Political Analysis},
  volume={32},
  number={4},
  pages={401--416},
  year={2024},
  publisher={Cambridge University Press}
}

@inproceedings{sociobench2025,
  title={SocioBench: Modeling Human Behavior in Sociological Surveys with Large Language Models},
  author={Wang, Jia and Zhao, Ziyu and Ni, Tingjuntao and Wei, Zhongyu},
  booktitle={Proceedings of the 2025 Conference on Empirical Methods in Natural Language Processing},
  pages={26268--26300},
  year={2025}
}

@inproceedings{cao2025,
  title={Specializing large language models to simulate survey response distributions for global populations},
  author={Cao, Yong and Liu, Haijiang and Arora, Arnav and Augenstein, Isabelle and R{\"o}ttger, Paul and Hershcovich, Daniel},
  booktitle={Proceedings of the 2025 Conference of the Nations of the Americas Chapter of the Association for Computational Linguistics: Human Language Technologies (Volume 1: Long Papers)},
  pages={3141--3154},
  year={2025}
}

@inproceedings{santurkar2023whose,
  title={Whose opinions do language models reflect?},
  author={Santurkar, Shibani and Durmus, Esin and Ladhak, Faisal and Lee, Cinoo and Liang, Percy and Hashimoto, Tatsunori},
  booktitle={International conference on machine learning},
  pages={29971--30004},
  year={2023},
  organization={PMLR}
}

@article{dominguezolmedo2024questioning,
  title={Questioning the survey responses of large language models},
  author={Dominguez-Olmedo, Ricardo and Hardt, Moritz and Mendler-D{\"u}nner, Celestine},
  journal={Advances in Neural Information Processing Systems},
  volume={37},
  pages={45850--45878},
  year={2024}
}

@inproceedings{aher2023using,
  title={Using large language models to simulate multiple humans and replicate human subject studies},
  author={Aher, Gati V and Arriaga, Rosa I and Kalai, Adam Tauman},
  booktitle={International conference on machine learning},
  pages={337--371},
  year={2023},
  organization={PMLR}
}

@techreport{horton2023homo,
  title={Large language models as simulated economic agents: What can we learn from homo silicus?},
  author={Horton, John J and Filippas, Apostolos and Manning, Benjamin S},
  year={2023},
  institution={National Bureau of Economic Research}
}

@article{dillion2023can,
  title={Can AI language models replace human participants?},
  author={Dillion, Danica and Tandon, Niket and Gu, Yuling and Gray, Kurt},
  journal={Trends in Cognitive Sciences},
  volume={27},
  number={7},
  pages={597--600},
  year={2023},
  publisher={Elsevier}
}

@inproceedings{hamalainen2023evaluating,
  title={Evaluating large language models in generating synthetic hci research data: a case study},
  author={H{\"a}m{\"a}l{\"a}inen, Perttu and Tavast, Mikke and Kunnari, Anton},
  booktitle={Proceedings of the 2023 CHI conference on human factors in computing systems},
  pages={1--19},
  year={2023}
}

@article{sarstedt2024silicon,
  title={Using large language models to generate silicon samples in consumer and marketing research: Challenges, opportunities, and guidelines},
  author={Sarstedt, Marko and Adler, Susanne J and Rau, Lea and Schmitt, Bernd},
  journal={Psychology \& Marketing},
  volume={41},
  number={6},
  pages={1254--1270},
  year={2024},
  publisher={Wiley Online Library}
}

@article{li2024frontiers,
  title={Frontiers: Determining the validity of large language models for automated perceptual analysis},
  author={Li, Peiyao and Castelo, Noah and Katona, Zsolt and Sarvary, Miklos},
  journal={Marketing Science},
  volume={43},
  number={2},
  pages={254--266},
  year={2024},
  publisher={INFORMS}
}

@article{brucks2025prompt,
  title={Prompt architecture induces methodological artifacts in large language models},
  author={Brucks, Melanie and Toubia, Olivier},
  journal={PloS one},
  volume={20},
  number={4},
  pages={e0319159},
  year={2025},
  publisher={Public Library of Science San Francisco, CA USA}
}

@article{gao2025caution,
  title={Take caution in using LLMs as human surrogates},
  author={Gao, Yuan and Lee, Dokyun and Burtch, Gordon and Fazelpour, Sina},
  journal={Proceedings of the National Academy of Sciences},
  volume={122},
  number={24},
  pages={e2501660122},
  year={2025},
  publisher={National Academy of Sciences}
}

@article{lin2025six,
  title={Six fallacies in substituting large language models for human participants},
  author={Lin, Zhicheng},
  journal={Advances in Methods and Practices in Psychological Science},
  volume={8},
  number={3},
  pages={25152459251357566},
  year={2025},
  publisher={Sage Publications Sage CA: Los Angeles, CA}
}

@inproceedings{park2022social,
  title={Social simulacra: Creating populated prototypes for social computing systems},
  author={Park, Joon Sung and Popowski, Lindsay and Cai, Carrie and Morris, Meredith Ringel and Liang, Percy and Bernstein, Michael S},
  booktitle={Proceedings of the 35th annual ACM symposium on user interface software and technology},
  pages={1--18},
  year={2022}
}

@inproceedings{park2023generative,
  title={Generative agents: Interactive simulacra of human behavior},
  author={Park, Joon Sung and O'Brien, Joseph and Cai, Carrie Jun and Morris, Meredith Ringel and Liang, Percy and Bernstein, Michael S},
  booktitle={Proceedings of the 36th annual acm symposium on user interface software and technology},
  pages={1--22},
  year={2023}
}

@article{ziems2024can,
  title={Can large language models transform computational social science?},
  author={Ziems, Caleb and Held, William and Shaikh, Omar and Chen, Jiaao and Zhang, Zhehao and Yang, Diyi},
  journal={Computational Linguistics},
  volume={50},
  number={1},
  pages={237--291},
  year={2024}
}

@article{mou2026from,
  title={From individual to society: A survey on social simulation driven by large language model-based agents},
  author={Mou, Xinyi and Ding, Xuanwen and He, Qi and Wang, Liang and Liang, Jingcong and Zhang, Xinnong and Sun, Libo and Lin, Jiayu and Zhou, Jie and Xuanjing, Huang and others},
  journal={ACM Computing Surveys},
  volume={58},
  number={11},
  pages={1--41},
  year={2026},
  publisher={ACM New York, NY}
}

@inproceedings{deshpande2023toxicity,
  title={Toxicity in chatgpt: Analyzing persona-assigned language models},
  author={Deshpande, Ameet and Murahari, Vishvak and Rajpurohit, Tanmay and Kalyan, Ashwin and Narasimhan, Karthik},
  booktitle={Findings of the association for computational linguistics: EMNLP 2023},
  pages={1236--1270},
  year={2023}
}

@article{li2025persona,
  title={Llm generated persona is a promise with a catch},
  author={Li, Leon and Chen, Haozhe and Namkoong, Hongseok and Peng, Tianyi},
  journal={Advances in Neural Information Processing Systems},
  volume={38},
  year={2026}
}

@article{lutz2025prompt,
  title={The prompt makes the person (a): A systematic evaluation of sociodemographic persona prompting for large language models},
  author={Lutz, Marlene and Sen, Indira and Ahnert, Georg and Rogers, Elisa and Strohmaier, Markus},
  journal={arXiv preprint arXiv:2507.16076},
  year={2025}
}

@article{liang2023holistic,
  title={Holistic Evaluation of Language Models},
  author={Liang, Percy and Bommasani, Rishi and Lee, Tony and Tsipras, Dimitris and Soylu, Dilara and Yasunaga, Michihiro and Zhang, Yian and Narayanan, Deepak and Wu, Yuhuai and Kumar, Ananya and others},
  journal={Transactions on Machine Learning Research},
  volume={2023},
  year={2023},
  publisher={Transactions on Machine Learning Research}
}

@inproceedings{ribeiro2020checklist,
  title={Beyond accuracy: Behavioral testing of NLP models with CheckList},
  author={Ribeiro, Marco Tulio and Wu, Tongshuang and Guestrin, Carlos and Singh, Sameer},
  booktitle={Proceedings of the 58th annual meeting of the association for computational linguistics},
  pages={4902--4912},
  year={2020}
}

@article{srivastava2023bigbench,
  title={Beyond the imitation game: Quantifying and extrapolating the capabilities of language models},
  author={Srivastava, Aarohi and Rastogi, Abhinav and Rao, Abhishek and Shoeb, Abu Awal Md and Abid, Abubakar and Fisch, Adam and Brown, Adam R and Santoro, Adam and Gupta, Aditya and Garriga-Alonso, Adri{\`a} and others},
  journal={Transactions on machine learning research},
  year={2023}
}

@inproceedings{mitchell2019modelcards,
  title={Model cards for model reporting},
  author={Mitchell, Margaret and Wu, Simone and Zaldivar, Andrew and Barnes, Parker and Vasserman, Lucy and Hutchinson, Ben and Spitzer, Elena and Raji, Inioluwa Deborah and Gebru, Timnit},
  booktitle={Proceedings of the conference on fairness, accountability, and transparency},
  pages={220--229},
  year={2019}
}

@article{gebru2021datasheets,
  title={Datasheets for datasets},
  author={Gebru, Timnit and Morgenstern, Jamie and Vecchione, Briana and Vaughan, Jennifer Wortman and Wallach, Hanna and Iii, Hal Daum{\'e} and Crawford, Kate},
  journal={Communications of the ACM},
  volume={64},
  number={12},
  pages={86--92},
  year={2021},
  publisher={ACM New York, NY, USA}
}

@article{caliskan2017semantics,
  title={Semantics derived automatically from language corpora contain human-like biases},
  author={Caliskan, Aylin and Bryson, Joanna J and Narayanan, Arvind},
  journal={Science},
  volume={356},
  number={6334},
  pages={183--186},
  year={2017},
  publisher={American Association for the Advancement of Science}
}

@article{bolukbasi2016man,
  title={Man is to computer programmer as woman is to homemaker? debiasing word embeddings},
  author={Bolukbasi, Tolga and Chang, Kai-Wei and Zou, James Y and Saligrama, Venkatesh and Kalai, Adam T},
  journal={Advances in neural information processing systems},
  volume={29},
  year={2016}
}

@inproceedings{buolamwini2018gender,
  title={Gender shades: Intersectional accuracy disparities in commercial gender classification},
  author={Buolamwini, Joy and Gebru, Timnit},
  booktitle={Conference on fairness, accountability and transparency},
  pages={77--91},
  year={2018},
  organization={PMLR}
}

@inproceedings{sheng2019woman,
  title={The woman worked as a babysitter: On biases in language generation},
  author={Sheng, Emily and Chang, Kai-Wei and Natarajan, Prem and Peng, Nanyun},
  booktitle={Proceedings of the 2019 conference on empirical methods in natural language processing and the 9th international joint conference on natural language processing (EMNLP-IJCNLP)},
  pages={3407--3412},
  year={2019}
}

@inproceedings{bender2021dangers,
  title={On the dangers of stochastic parrots: Can language models be too big?},
  author={Bender, Emily M and Gebru, Timnit and McMillan-Major, Angelina and Shmitchell, Shmargaret},
  booktitle={Proceedings of the 2021 ACM conference on fairness, accountability, and transparency},
  pages={610--623},
  year={2021}
}

@inproceedings{parrish2022bbq,
  title={BBQ: A hand-built bias benchmark for question answering},
  author={Parrish, Alicia and Chen, Angelica and Nangia, Nikita and Padmakumar, Vishakh and Phang, Jason and Thompson, Jana and Htut, Phu Mon and Bowman, Samuel R},
  booktitle={Findings of the Association for Computational Linguistics: ACL 2022},
  pages={2086--2105},
  year={2022}
}

@article{hardt2016equality,
  title={Equality of opportunity in supervised learning},
  author={Hardt, Moritz and Price, Eric and Srebro, Nati},
  journal={Advances in neural information processing systems},
  volume={29},
  year={2016}
}

@inproceedings{selbst2019fairness,
  title={Fairness and abstraction in sociotechnical systems},
  author={Selbst, Andrew D and Boyd, Danah and Friedler, Sorelle A and Venkatasubramanian, Suresh and Vertesi, Janet},
  booktitle={Proceedings of the conference on fairness, accountability, and transparency},
  pages={59--68},
  year={2019}
}

@article{shmueli2011predictive,
  title={Predictive analytics in information systems research1},
  author={Shmueli, Galit and Koppius, Otto R},
  journal={MIS quarterly},
  volume={35},
  number={3},
  pages={553--572},
  year={2011},
  publisher={Management Information Systems Research Center, University of Minnesota}
}

@article{arnott2005critical,
  title={A critical analysis of decision support systems research},
  author={Arnott, David and Pervan, Graham},
  journal={Journal of information technology},
  volume={20},
  number={2},
  pages={67--87},
  year={2005},
  publisher={SAGE Publications Sage UK: London, England}
}

@article{saibene2021expert,
  title={Expert systems: Definitions, advantages and issues in medical field applications},
  author={Saibene, Aurora and Assale, Michela and Giltri, Marta},
  journal={Expert Systems with Applications},
  volume={177},
  pages={114900},
  year={2021},
  publisher={Elsevier}
}

@article{okeefe1993expert,
  title={Expert system verification and validation: a survey and tutorial},
  author={O'Keefe, Robert M and O'Leary, Daniel E},
  journal={Artificial intelligence review},
  volume={7},
  number={1},
  pages={3--42},
  year={1993},
  publisher={Springer}
}

@article{groves2010total,
  title={Total survey error: Past, present, and future},
  author={Groves, Robert M and Lyberg, Lars},
  journal={Public opinion quarterly},
  volume={74},
  number={5},
  pages={849--879},
  year={2010},
  publisher={Oxford University Press}
}

\end{document}